\documentclass[journal]{IEEEtran}

\usepackage{graphicx}
\usepackage{array}
\usepackage{makecell}
\usepackage{multirow}
\usepackage{amsmath,amsfonts}
\usepackage[normalem]{ulem}
\usepackage{tcolorbox}
\usepackage{colortbl}
\usepackage{color}
\usepackage{enumitem}
\usepackage{url}
\usepackage{booktabs}
\usepackage{xspace}

\usepackage{threeparttable}
\usepackage{ulem}
\usepackage{amssymb}

\usepackage{float}
\usepackage{balance}
\usepackage[caption=false,font=normalsize,labelfont=sf,textfont=sf]{subfig}
\usepackage{textcomp}
\usepackage{stfloats}
\usepackage{verbatim}
\usepackage{balance}
\usepackage{hyperref}
\usepackage{makecell}
\usepackage{multirow}
\usepackage{graphicx}
\usepackage{amsmath}
\usepackage{color}
\usepackage{colortbl}
\usepackage{CJKutf8}
\usepackage{setspace}
\usepackage{algorithm, algorithmicx, algpseudocode}
\usepackage{comment}
\begin{document}

\title{ARIES: Relation Assessment and Model Recommendation for Deep Time Series Forecasting}

\author{Fei Wang, Yujie Li, Zezhi Shao, Chengqing Yu, Yisong Fu, Zhulin An, Yongjun Xu, Xueqi Cheng
\thanks{Fei Wang, Yujie Li, Zezhi Shao, Chengqing Yu, Yisong Fu, Zhulin An, Yongjun Xu, Xueqi Cheng are with the Institute of Computing Technology, CAS, Beijing 100190, China.
Fei Wang, Yujie Li, Chengqing Yu, Yisong Fu, Zhulin An, Yongjun Xu, Xueqi Cheng are also with the University of Chinese Academy of Sciences, Beijing 100049, China. (e-mail: \{wangfei, liyujie23s, shaozezhi, yuchengqing22b, fuyisong24s, anzhulin, xyj, cxq\}@ict.ac.cn)
}
\thanks{Fei Wang and Yujie Li contribute equally to this work and should be considered co-first authors.}
\thanks{Corresponding Authours: Zezhi Shao (shaozezhi@ict.ac.cn) and Xueqi Cheng (cxq@ict.ac.cn).}}

{}


\maketitle

\begin{abstract}
Recent advancements in deep learning models for time series forecasting have been significant. These models often leverage fundamental time series properties such as seasonality and non-stationarity, which may suggest an intrinsic link between model performance and data properties. However, existing benchmark datasets fail to offer diverse and well-defined temporal patterns, restricting the systematic evaluation of such connections. Additionally, there is no effective model recommendation approach, leading to high  time and cost expenditures when testing different architectures across different downstream applications.
For those reasons, we propose ARIES, a framework for assessing relation between time series properties and modeling strategies, and for recommending deep forcasting models for realistic time series. First, we construct a synthetic dataset with multiple distinct patterns, and design a comprehensive system to compute the properties of time series. Next, we conduct an extensive benchmarking of over 50 forecasting models, and establish the relationship between time series properties and modeling strategies. Our experimental results reveal a clear correlation. Based on these findings, we propose the first deep forecasting model recommender, capable of providing interpretable suggestions for real-world time series.
In summary, ARIES is the first study to establish the relations between the properties of time series data and modeling strategies, while also implementing a model recommendation system. The code is available at: \url{https://github.com/blisky-li/ARIES}.
\end{abstract}


\section{Introduction}\label{intro}

\IEEEPARstart{D}{eep} Time Series Forecasting~(DTSF) aims to predict future observations by capturing the underlying temporal patterns through deep learning models. The fascination with predicting the future has driven significant research interest, with applications in fields such as  finance~\cite{fang2024spatio}, climate science~\cite{huang2023extreme}, transportation~\cite{li2024dynamic, li2025sta}, and healthcare~\cite{germainshape}.

Real-world time series are heterogeneous due to domain-specific characteristics, each exhibiting distinct statistical properties.  For example, electricity and climate time series show strong seasonality due to natural cycles, while financial time series often exhibit relative stationarity, reflecting the temporal patterns of different systems. Recently, heterogeneity has been identified as a central challenge~\cite{shao2023exploring, qiu2024tfb}  in deep time series forecasting,  requiring extensive testing and expert experience to select and design the appropriate models for downstream applications.

\begin{figure*}[t]
	
	\centering
	\includegraphics[width=1\textwidth]{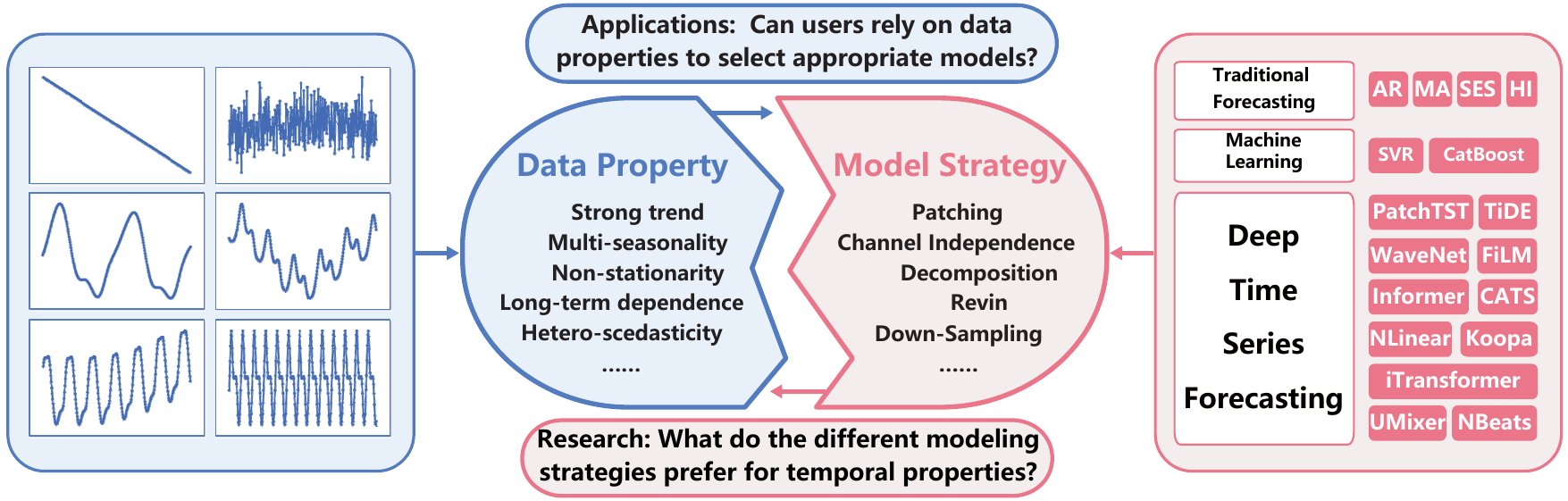}
	\caption{Possible potential relation between time series properties and modeling strategies. 
	Time series~(left, blue) across domains like healthcare, transport, and energy exhibit diverse patterns~(properties) due to system-specific dynamics. Among forecasting techniques~(right, red), traditional models rely on explicit property analysis, modern deep models follow a black-box approach, often tailoring various modeling strategies to fit certain properties without prior analysis. However, there is no clear framework to assess whether modeling strategies align with specific properties, and deep forecasting is hard to adopt in practical applications due to the need for extensive model trial-and-error and the lack of interpretability. ARIES bridges this gap by revealing the link between properties and modeling strategies, offering guidance for further research and providing reliable forecasting model recommendation tools with explanations to industry.}
	\label{ddmt}
\end{figure*}

Furthermore, despite the variety of architectures~\cite{shao2023exploring,miller2024survey,wang2024tssurvey}, DTSF models often share similar modeling strategies due to the common nature of time series properties. Typical combinations include series decomposition~\cite{ wu2021autoformer,zhou2022fedformer,woo2022etsformer,wangtimemixer,cleveland1990stl} with trend and seasonality, reversible instance normalization~\cite{kim2021reversible, zhou2022film,liu2022non,yu2023dsformer,liuitransformer} with hetero-scedasticity, patching~\cite{nietime,linsparsetsf, lin2023segrnn,ma2024u} with memorability and so on.

Recent studies have shown that no single deep model achieves state-of-the-art performance across all time series~\cite{brigato2025position, wang2024deep}. In addition, BasicTS~\cite{shao2023exploring} highlights significant variations in performance across datasets, and Monash~\cite{godahewa2021monash} finds simple methods sometimes outperform early deep learning techniques. These findings raise the following questions~(see Figure~\ref{ddmt}): \textit{(1) How do deep modeling strategies favor specific time series properties? (2) How to recommend deep models based on time series properties?} While some benchmarking efforts~\cite{shao2023exploring,godahewa2021monash,qiu2024tfb,zhang2023probts, aksu2024giftevalbenchmarkgeneraltime,li2024foundts} have explored advancements in time series forecasting and provided fair comparisons, they still fall short of achieving these two goals. A comparison of these benchmarks with the approach in this paper is shown in Table~\ref{benchmark}.

Specifically, existing benchmarking datasets suffer from limited pattern coverage and uncontrollability, which hinders the effective assessment of the relations between data properties and modeling strategies. Domain-specific datasets often capture only partial temporal patterns. For instance, the electricity dataset focuses solely on seasonality~\cite{qiu2024tfb}, while it remains debatable whether the exchange rate dataset contains learnable patterns ~\cite{shao2023exploring}. Moreover, uncontrollable distribution shifts ~\cite{kim2021reversible} disrupt temporal continuity and undermine the assumption that historical patterns can generalize to future observations, leading to unreliable evaluation results.

Furthermore, due to inadequate relation assessment in these benchmarks, the recommendation of deep forecasting models has been neglected. In practical scenarios, superior forecasting performance alone is not enough; the process must also be interpretable to support informed decision-making and reduce trial-and-error costs. However, existing benchmarks, such as TFB~\cite{qiu2024tfb} and BasicTS~\cite{shao2023exploring}, provide only vague guidance, lacking a systematic recommendation framework, and offering no strategic advice or interpretability rationals.

\begin{table}[t]
	\centering 		 		 		
	\caption{Comparison of existing time series benchmarks.}
	\small
	\label{benchmark}
	\renewcommand\arraystretch{0.7}
	\resizebox{1\linewidth}{!}{

		\begin{tabular}{c|c|c|c}
			\toprule[1.5pt]
			Benchmark & \makecell[c]{Controllable Datasets \\ with Diverse Patterns} & \makecell[c]{Relation \\ Assessment} & \makecell[c]{Model \\ Recommendation} \\
			\midrule
			BasicTS~\cite{shao2023exploring}& $\bigcirc$&$\bigcirc$&$\times$\\
			TSlib~\cite{wutimesnet} &$\bigcirc$&$\times$&$\times$\\
			Monash~\cite{godahewa2021monash} &$\bigcirc$&$\times$&$\times$\\
			TFB~\cite{qiu2024tfb}&$\bigcirc$&$\bigcirc$&$\times$\\
			ProbTS~\cite{zhang2023probts}&$\bigcirc$&$\bigcirc$&$\times$\\
			GIFT-Eval~\cite{aksu2024giftevalbenchmarkgeneraltime} &$\bigcirc$&$\bigcirc$&$\times$\\
			FoundTS~\cite{li2024foundts} &$\bigcirc$&$\times$&$\times$\\
			\midrule
			ARIES &\checkmark&\checkmark& \checkmark\\
			\bottomrule[1.5pt]
			\multicolumn{4}{c}{\footnotemark{$\times$ indicates absent, \checkmark
					indicates present, $\bigcirc$ indicates incomplete.}}
		\end{tabular}
	}
\end{table}

As a result, we propose \textbf{ARIES}, a novel framework for relation \textbf{A}ssessment and model \textbf{R}ecommen-dation for deep time ser\textbf{IES} forecasting: (i) Synthetic dataset ~\cite{ansari2024chronos, dooley2023forecastpfn} with diverse patterns named \textit{Synth} and comprehensive time series property evaluation system; (ii) Benchmark called \textit{ARIES TEST} and research on relation assessment between time series properties and modeling strategies based on forecasting performance of 50+ baselines; (iii) Recommendation framework for deep forecasting that provides appropriate models, strategy preferences and interpretable insights.
In summary, Our contributions are as follows:

\begin{itemize}
	\item To comprehensively and reliably analyze forecasting models, we synthesize a controllable time series dataset \textit{Synth} with diverse patterns by Gaussian Process.
	\item Based on \textit{Synth} and property evaluation system, we assess the relation between time series properties and modeling strategies on 50+ baselines, and propose relevant benchmarks called \textit{ARIES TEST}.
	\item We implement the first interpretable recommendation framework for deep time series forecasting to recommend appropriate models to realistic time series.
\end{itemize}

The paper is organized as follows: Section~\ref{related} reviews related work and Section~\ref{prepara} details the preparation of ARIES. Section~\ref{relation} provides relation assessment through 
experiments and findings. Section~\ref{model} describes the model recommendation system and its experimental analysis.  Appendix~E\,\&\,F discuss the limitations and future work of ARIES.

\section{Related Work}\label{related}

\subsection{Time Series Property}
Statistical properties are fundamental to time series analysis. For example, stationarity~\cite{box1968some} requires  constant mean, variance, and autocorrelation, trend~\cite{woo2022etsformer} captures macro-level changes such as growth or decline. Long short-term dependencies~\cite{zhou2021informer} quantify memorability in deep learning contexts. Moreover, properties are measurable through statistical tests, such as Augmented Dickey-Fuller~(ADF)~\cite{dickey1979distribution} Test for stationarity, Lagrange Multiplier~(LM) for ARCH~\cite{akgiray1989conditional} effects.  ARIES fully integrates properties and their computational metrics as analytical tools.
\vspace{-5pt}
\subsection{Time Series Synthesis}
Recent advancements in time series synthesis have stemmed from foundational models that leverage high-quality synthetic data to augment training datasets. Notable works include Fourier-transform-based sinusoidal generation in Moment~\cite{goswami2024moment} and TimeFM~\cite{das2023decoder}, STL-like combinations of seasonal, trending, and noise components in ForecastPFN~\cite{dooley2023forecastpfn}, and Gaussian kernel processes in Chronos~\cite{ansari2024chronos}. ARIES adopts the process-controllable synthesis method to generate the \textit{Synth} dataset with diverse and stable patterns.
\vspace{-5pt}
\subsection{Time Series Forecasting Models}
Traditional forecasting methods are based on mathematical models and rely on property testing for model selection or parameter tuning. ARMA and ARIMA~\cite{box1968some,hyndman2008forecasting} require stationarity verification via ADF tests, and ETS requires predefined seasonal lengths.

Deep forecasting, though considered a black box, still maintains property specialization. For example, FEDformer~\cite{zhou2022fedformer}  enhances seasonal learning through frequency domain, NSTransformer~\cite{liu2022non} captures non-stationarity, and PatchTST~\cite{nietime} addresses long short-term dependencies with Patch strategy. Additionally, channel strategies and debates on Transformer vs. MLP~\cite{zeng2023transformers} are also related to specific properties.
\vspace{-5pt}
\subsection{Time Series Forecasting Benchmark}\label{tsfb}
Existing benchmarks have driven the development of time series forecasting through unified pipelines and innovative insights. However, as demonstrated in Section~\ref{intro} and Table~\ref{benchmark}, current benchmarks face problems in exploring the link between time series properties and modeling strategies due to uncontrollable benchmark datasets with pattern scarcity. These factors undermine both analytical reliability and practical applicability.

\begin{figure*}[htb]
	\centering
	\includegraphics[width=1\textwidth]{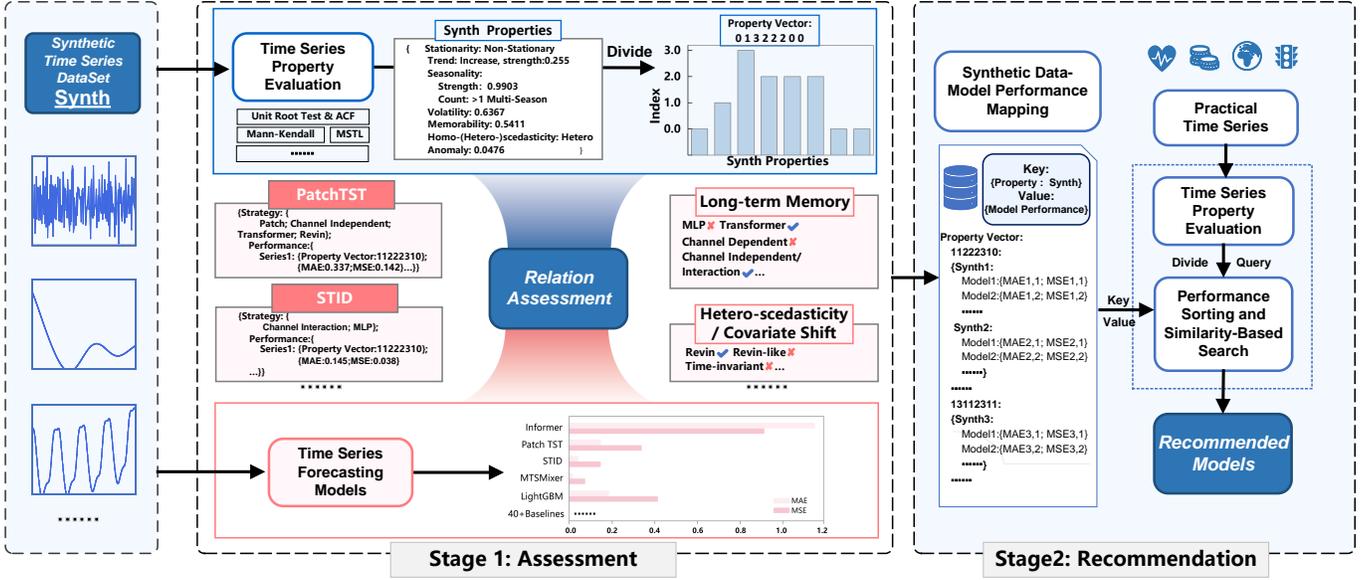}
	\caption{ARIES: Relation Assessment and Model Recommendation based on Time Series Properties and Performance}
	\label{aries}
\end{figure*}

\section{Preparation of ARIES}\label{prepara}

At first, we present a brief overview of the ARIES as shown in Figure~\ref{aries}:

\textit{\textbf{Assessment}}: 

\begin{itemize}
	\item[1] Construct synthetic dataset \textit{Synth} with diverse controllable patterns~(Blue part on left);
	\item[2] Compute the properties of historical part of each series in \textit{Synth}~(Top center blue part);
	\item[3] Summarize modeling strategies behind 50+ baselines and record their forecasting performance on \textit{Synth}~(Bottom center red part);
	\item[4] With the above benchmark \textit{ARIES TEST}, assess the relation between time series properties and modeling strategies~(Center).
\end{itemize}

\textit{\textbf{Recommendation}}~(Right part): 
\begin{itemize}
	\item[1] Establish mappings between properties of each series in \textit{Synth} and model performances.
	\item[2] Calculate practical dataset properties like the step 2 of \textit{Assessment}.
	\item[3] Match similar \textit{Synth} series by properties and record their model performances.
	\item[4] Generate recommended deep forecasting models and provide transparent suggestions.
\end{itemize}

\vspace{5pt}
This section delineates the methodological foundations of ARIES through three key preparations: time series property evaluation, synthetic data generation, and forecasting model strategy analysis.

\subsection{Time Series Properties}\label{etsp}

\subsubsection{Selected properties and visualizations}
We systematically define seven critical time series properties, selection rationale, quantitative evaluation metrics and visualization of typical samples, including Stationarity, Trend, Seasonality, Volatility, Memorability, Homo-(Hetero-)scedasticity and Anomaly.

\textbf{Stationarity} requires constant mean, variance, and auto-correlation over time. 
Stationarity~\cite{box1968some,hyndman2008forecasting} marks the start of financial and mathematical time series analysis, and non-stationarity is the original motivation for deep forecasting~\cite{zhou2021informer}.

ARIES adopts strictly-sense stationarity~(SSS) as the criterion, including unit root tests by Augmented Dickey-Fuller~(ADF) and Kwiatkowski-Phillips-Schmidt-Shin~(KPSS) for wide-sense stationarity~(WSS), as well as autocorrelation-convergent~ACF determination:
\begin{flalign}
	\text{Is\_Stationary} = \left\{
	\begin{aligned}
		&1 \,\,\text{ if } ADF(x) \ \text{and } KPSS(x) \\
		&\hspace{7.0em} \text{and } ACF(x) \\
		&0 \,\,\text{ else} \\
	\end{aligned}
	\right.
\end{flalign}
Typical strictly-sense stationary time series such as white noise are shown in Figure~\ref{stationary}:

\begin{figure}[ht]
	\centering
	\includegraphics[width=0.33\textwidth]{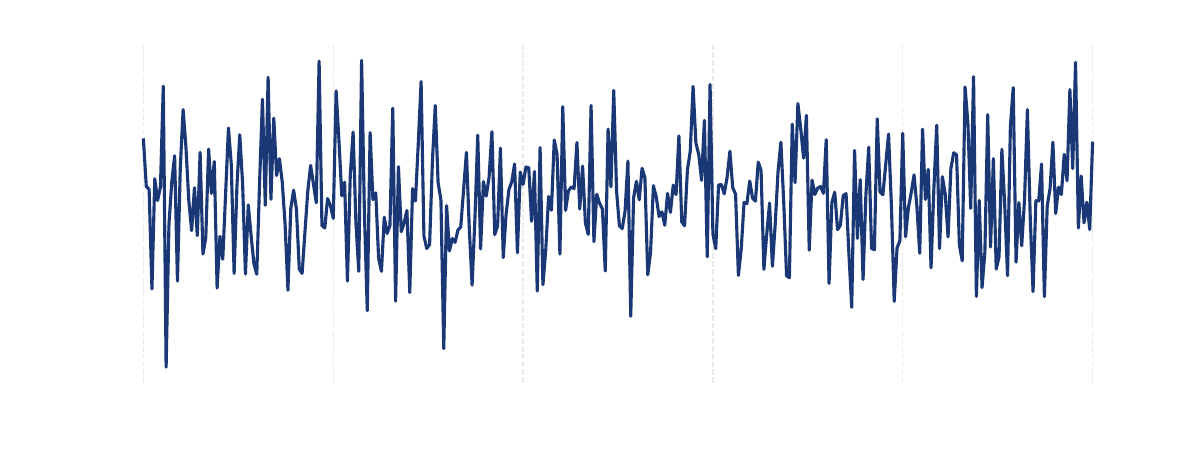}
	\caption{Strictly stationary time series. This implies the absence of learnable patterns like trend or seasonality. Exception: straight lines with no trend—unrealistic and of little research value.}
	\label{stationary}
\end{figure}

\textbf{Trend} quantifies directional patterns in time series through upward, downward, or stable directional changes, serving as a fundamental component in seasonal-trend decomposition~(STL)~\cite{cleveland1990stl, bandara2021mstl} and core motivations for forecasting~\cite{hyndman2008forecasting, woo2022etsformer}. ARIES implements the Mann-Kendall test to compute trend magnitude~($Trend\_Strength$) with strength values bounded in [-1,1]:
\begin{flalign}
	Trend\_Strength = Mann\_Kendall(x)
\end{flalign}

\begin{figure*}[t]
	\centering
	\includegraphics[width=1\textwidth]{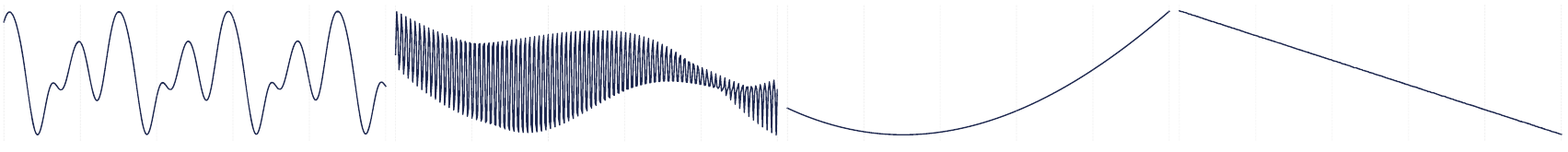}
	\caption{Time series with increasing trend strength. ARIES does not distinguish upward or downward trends, as trend strength is symmetric. Typically, trend and seasonality are mutually exclusive;  when coupled, moderate trend strength is assumed to indicate challenging forecasting.}
	\label{trend}
\end{figure*}

As shown in Figure~\ref{trend}, from a seasonal series with no trend to a completely up-or-down straight line, the time series trend strength gradually increases and is usually mutually exclusive with the season.
\begin{figure*}[t]
	\centering
	\includegraphics[width=1\textwidth]{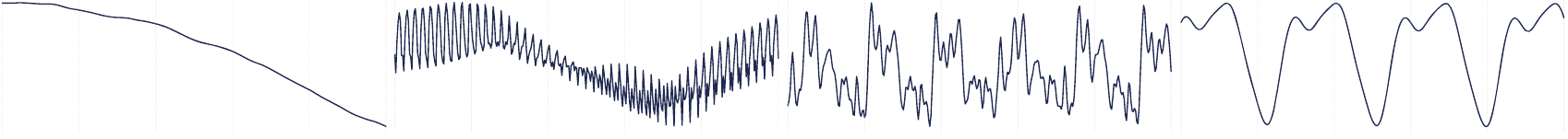}
	\caption{Time series with increasing season strength. Trend strength can also be computed using the seasonality formula, but Mann-Kendall avoids the information loss caused by multi-season detection.}
	\label{seasonstrength}
\end{figure*}

\begin{figure*}[t]
	\centering
	\includegraphics[width=0.75\textwidth]{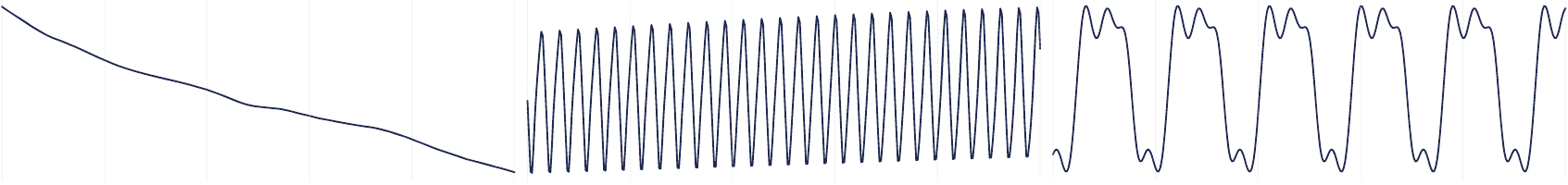}
	\caption{Time series with increasing season counts.}
	\label{seasoncount}
\end{figure*}
\textbf{Seasonality} and periodicity are not distinguished in ARIES and are both defined as recurring temporal patterns. Seasonality is likewise a central part of STL and motivates a huge amount of work~\cite{zhou2022fedformer, piao2024fredformer, lin2024cyclenet, zhou2022film, yi2024frequency}. Calculating seasonal strength requires precise season identification prior to quantification, presenting greater complexity than trend analysis. ARIES adopts an auto-correlation function-based approach shown in Algorithm.~\ref{alg:msd} to detect $Top\mbox{-} K$ primary seasons.

\begin{algorithm}[ht]
	\setstretch{1}
	\renewcommand{\algorithmicrequire}{\textbf{Input:}}
	\renewcommand{\algorithmicensure}{\textbf{Output:}}
	\caption{Multi-season Detection}
	\label{alg:msd}
	
	\begin{algorithmic}[1]
		
		\Require
		\Statex Time series $\mathbf{X} = \{x_i\}_{i=1}^L$ where $L$ is the length; 
		\Statex Max candidate seasons $K \in \mathbb{N}^+$ (default: 10);
		\Statex Sampling frequency $f_s \in \mathbb{R}^+$ (default: 1.0);
		
		\Ensure  Collection of primary seasons $\mathcal{P}_{primary}$, where $p \in \mathcal{P}_{primary}$ and $p < L//2$;
		
		\State Remove trend components with first-order differencing $\mathbf{X} \gets \{x_i - x_{i-1}\}_{i=2}^L$
		
		\State Compute autocorrelation function (ACF)
		$\quad \mathbf{R} \gets \text{ACF}(\mathbf{X}, \text{lag}=L//2)$ 
		
		\State Identify candidate lags $\quad \mathcal{L}_{\text{raw}} \gets \text{argsort}(\mathbf{R})[1:]$
		
		\State Initialize candidate set $\mathcal{P}_{\text{cand}} \gets \emptyset$		 
		
		\For {$k \in \mathcal{L}_{\text{raw}}$}
		\State Compute season: $p \gets \lfloor k/f_s \rfloor$
		\If {$p \geq 2$ \textbf{and} $\mathbf{R}(p) > 0.1$ \textbf{and} $\forall p_i \in \mathcal{P}_{\text{cand}} \ \text{s.t.} \ p \bmod p_i \ge 2$} 
		
		\State	$\mathcal{P}_{\text{cand}} \gets \mathcal{P}_{\text{cand}} \cup \{p\}$
		\EndIf
		\EndFor

		\State Sort candidates: $\mathcal{P}_{\text{cand}} \gets \text{Sort}(\mathcal{P}_{\text{cand}})$
	\end{algorithmic}
	
\end{algorithm}

Multi-Seasonal and Trend-Like Decomposition using LOESS~(MSTL)~\cite{cleveland1990stl, bandara2021mstl} is used to decompose time series into residual ($R$), trend ($T$), and multi-seasonal ($\{S_{i}\}$) components. 
Seasonal strength across multiple seasons is then computed, where $var$ represents the variance.

\begin{flalign}
	[S_{0},...,S_{topk}], T, R = MSTL(x, [period_{0},..., period_{topk}])
\end{flalign}
\vspace{-10pt}
\begin{flalign}
	Season\_Strength = max(0, 1 - \frac{var(R)}{var(R + \sum_{i}^{topk}S_{i})} )
\end{flalign}
Time series progressively excludes non-repeating patterns such as trends or noise as season strength increases in Figure~\ref{seasonstrength}, and count increasing in Figure~\ref{seasoncount} also implies more complex repeating patterns.

\begin{figure*}[t]
	\centering
	\includegraphics[width=\textwidth]{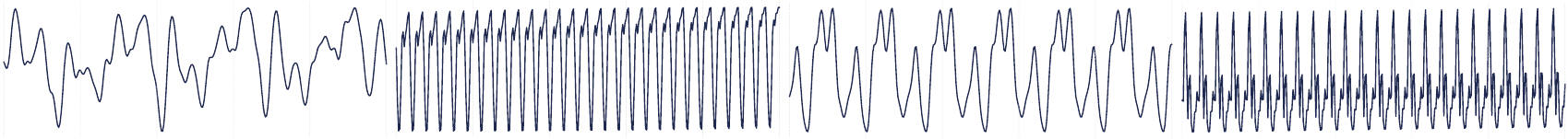}
	\caption{Time series with increasing volatility. Increased volatility implies larger value changes between timestamps, challenging the model’s numerical representation and fitting capacity.}
	\label{volatility}
\end{figure*}

\textbf{Volatility} measures the magnitude of temporal variations. It directly affects value modeling in deep forecasting, fluctuation space for probabilistic forecasting ~\cite{salinas2020deepar, woo2024moirai}, and token strategy of foundation model ~\cite{ansari2024chronos}, which provides support for broad time series analysis. For cross-scale comparisons, we utilized the coefficient of variation (CV) of the response standard deviation $\sigma$ relative to the mean $\mu$:
\begin{flalign}
	CV = \frac{\sqrt{\frac{1}{L}\sum_{i=1}^{L}(x_{i}^{'} - \mu)^2}}{\mu}
\end{flalign}
As qualifying the seasonal series in Figure~\ref{volatility}, an increase in volatility means more drastic changes between timestamps.
\begin{figure*}[t]
	\centering
	\includegraphics[width=\textwidth]{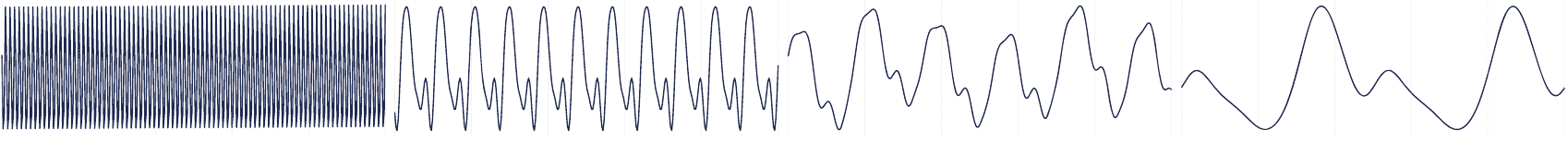}
	\caption{Time series with increasing memory. In addition to strong trends indicating long-term dependency, longer periods also imply longer dependencies. However, memory is a relative concept, which is dependent on season's length and window size.}
	\label{memory}
\end{figure*}
\begin{figure*}[t]
	\centering
	\includegraphics[width=0.48\textwidth]{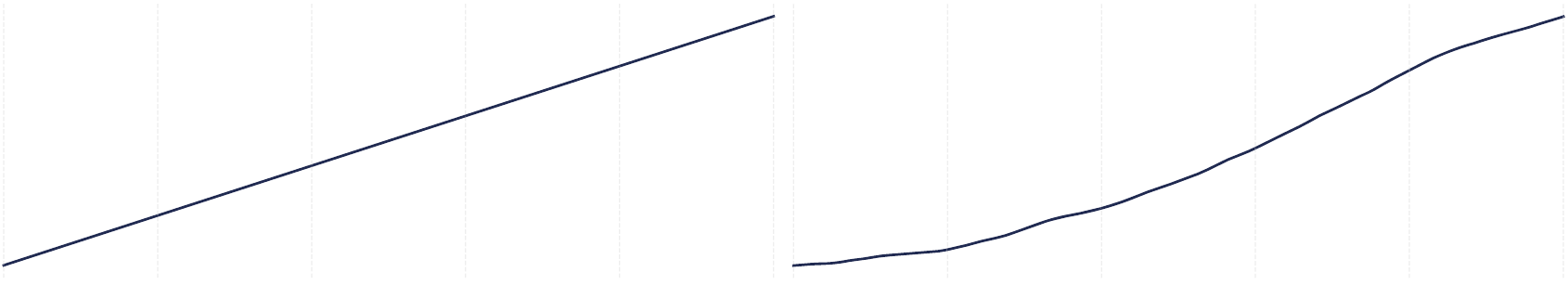}
	\caption{Homo-scadasticity and Hetro-scadasticity time series. Statistically, homo-scedasticity is strictly defined, so even slight variations imply hetero-scedasticity, which is common in real-world data.}
	\label{scadasticity}
\end{figure*}
\begin{figure*}[t]
	\centering
	\includegraphics[width=\textwidth]{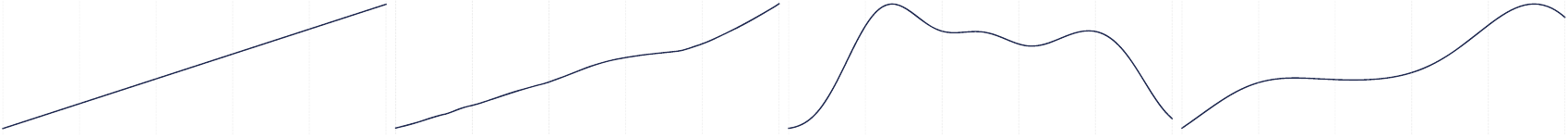}
	\caption{Time series with increasing anomaly. Z-score detection captures not only noise but also overall pattern shifts. Note that volatility reflects discrete changes between timestamps, while large anomalies indicate continuous value deviations.}
	\label{Anomaly}
\end{figure*}

\textbf{Memorability} quantifies the persistence of temporal dependencies, reflecting how historical values influence future states. The analysis of long and short-term dependencies has a well-established history in deep learning and is an extremely important motivation in deep temporal forecasting~\cite{zhou2021informer,wu2021autoformer, zhou2022fedformer, nietime, zhou2022film, daslong, linsparsetsf, lin2023segrnn}
ARIES adopts the Hurst exponent for this measure,  with a change from 0 to 1 pressing longer-term dependency.
\begin{flalign}
	Memory = Hurst(x)
\end{flalign}
As qualifying the seasonal series in Figure~\ref{memory}, memorability or long-term dependence of time series increases when the length of seasons increases, which will raise modeling difficulty.

\textbf{Homo-(Hetero-)scedasticity} characterizes temporal stability of variance, which is closely related to covariate shifts~\cite{kim2021reversible},and comes from Engle's ARCH framework~\cite{engle1982autoregressive} for addressing the constant variance assumption.
We follow the ARCH convention of judging scedasticity by applying Lagrange multiplier~(LM) test on series residuals ~($R$):
\begin{flalign}
	Scedasticity = \left\{
	\begin{aligned}
		&Homo \,\,\,\,\text{ if }  LM\_test(R) > 0.05 \\ 
		&Hetero \,\,\text{ else } &  \\ 
	\end{aligned}\right.
\end{flalign}

Despite the strong trend series all over Figure~\ref{scadasticity}, local fluctuations lead to hetero-scedasticity, which is extremely common in reality and makes forecasting difficult.

\textbf{Anomaly} describes observations that deviate significantly from others. Low anomaly scores indicate random noise, while high values are often associated with mean shift~\cite{kim2021reversible}. Using z-scores with 95\% confidence threshold~(1.645 one-tailed), we calculate anomaly as the proportion of outliers:
\begin{flalign}
	Anomaly = \frac{|\{x_{i} \in x | \frac{x_{i} - \mu}{\sigma} > 1.645\}|}{L}
\end{flalign}
As qualifying the non-seasonal series in Figure~\ref{Anomaly}, an increase in anomalies implies an evolution from random perturbations to mean shift. Mean shift represented by high anomalies, together with hetero-scedasticity, is the topic of time series distribution shift~\cite{kim2021reversible}.

Further discussions on property selection rationale and stability validation are provided in Appendix A.

\subsection{\textbf{Synthetic Time Series Datasets}}
\subsubsection{Methodology}
As mentioned earlier, the limitations of model variety, the lack of a priori knowledge, and the pitfalls of distributional drift prevent comprehensive and accurate relation assessments through benchmark datasets. Inspired by time series foundation models~\cite{dooley2023forecastpfn, goswami2024moment, das2023decoder}, especially Chronos~\cite{ansari2024chronos}, we chose to adopt the Gaussian process to synthesize the time series for subsequent in-depth analysis.

Gaussian processes (GPs), proposed as early as 1940s, are classical probabilistic models that define distributions over functions using a mean function \( m(t) \) and a positive definite kernel \( k(t, t') \). The kernel acts as a covariance function, governing relationships between function values across inputs. Integrated into modern libraries like \texttt{sklearn}\footnote{https://scikit-learn.org/stable/modules/gaussian\_process.html}, GP regression remains a foundational tool for regression tasks~\cite{seeger2004gaussian}.

\begin{enumerate}
	\item To generate synthetic time series covering diverse temporal patterns, we adopt all the kernel functions from \texttt{sklearn} and set reasonable parameters to form our kernel bank \( \mathcal{K} \).
	
	\item The final composite kernel \( k(t, t') \) is generated by:
	\begin{itemize}
		\item Randomly sampling \( j \sim \mathcal{U}(1, J) \) (where \( J = 3 \)) kernels from \( \mathcal{K} \)
		\item Combining them via binary operations (\( + \) or \( * \))
	\end{itemize}
	
	\item A synthetic time series of length \( l_{\text{syn}} \) is sampled from the GP prior
	$\mathcal{GP}\big(0,\, k(t, t')\big)$.
\end{enumerate}
The mathematical restrictions of the Gaussian process are proved and illustrated in Appendix B.

\subsubsection{Details and Advantages}

In ARIES, we constrain composite series to a maximum of three base kernel functions and incorporate Matern kernels to introduce uncertainty. This process generates \textit{Synth}, a synthetic time series dataset consisting of 1,500 sequences, each with 8,192 timestamps, which achieves the aforementioned benefits and is larger than most benchmark datasets. 

Gaussian kernel processes are sufficient to cover Fourier-based methods in Moment~\cite{goswami2024moment} and TimeFM~\cite{das2023decoder} and STL-like method in ForecastPFN~\cite{dooley2023forecastpfn}. By combining simple kernels into complex temporal patterns, it ensures \textit{Synth}'s advantages over Benchmark datasets:
\begin{itemize}
	\item \textbf{Diverse Patterns}: The great advantage of synthetic data is that it can obtain various time series properties by configuring different Gaussian kernels.
	\item \textbf{Stable Control}: Kernel process ensures consistent patterns throughout the series and specific kernels generate predefined properties, such as sinusoidal kernels for seasonality.
	\item \textbf{Fine-Grained Analysis:}  Kernel parameter adjustments facilitate a nuanced representation of properties, allowing for detailed and precise property analysis.
	\item \textbf{Distinctive Patterns: } Clear patterns are not too complex, cross-validation with a priori kernel combinations, property calculations, and manual observations enable researchers to quickly understand individual sequences.
\end{itemize}

\begin{figure}[t]
	
	\centering
	\includegraphics[width=0.48\textwidth]{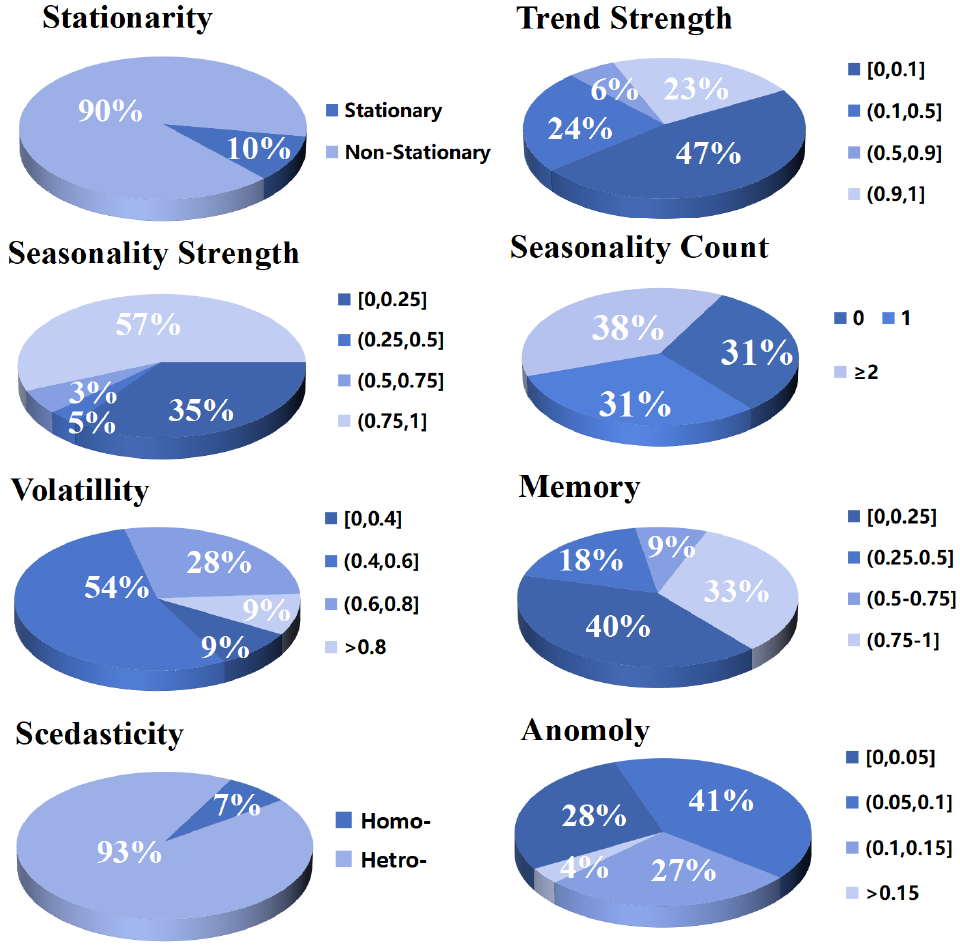}
	\caption{Distribution of data patterns. ARIES ensures sufficient quantity for each pattern type, bur distribution imbalance is inevitable~\cite{shao2025blast}. After excluding stationary series, 1\% of the \textit{Synth} still yields 16K training samples, which is typically sufficient for effective deep learning.}
	\label{datasetsbing}
\end{figure}

Moreover, we analyze the historical part of test set in \textit{Synth} with our time series properties evaluation system,  and divide property intervals as illustrated in Figure~\ref{datasetsbing}. Stationarity and scedasticity receive binary labels, seasonality count is categorized as non-seasonal~(0), single-season~(1), or multi-season~($\ge 2$), and other properties are partitioned into weak, moderate, strong, or unique intervals based on empirical value distributions.

\begin{table*}[t]
	\centering 		 		 		
	\caption{Initial relation on modeling strategy and time series properties.}
	\renewcommand\arraystretch{0.85}
	\label{ARIES_M} 	
	\resizebox{1\linewidth}{!}{
		\begin{tabular}{c|l}
			\toprule[1.5pt]
			\hline
			\small \textbf{Data Property}&\small \textbf{Modeling Strategy} \\
			\midrule
			\hline
			\multirow{6}{*}{\small \textbf{Stationarity}}&\textbf{Stationary}: AR, MA, ARMA, HI\\ 
			&\textbf{Non-stationary}: ARIMA, Koopa, NSformer,UMixer\\
			&\makecell[l]{\textbf{RevIN}: CATS, CycleNet, DSformer, FiLM, iTransformer, Koopa, MTSMixer, PatchTST, SOFTS, Sumba,\\ \quad\quad\quad TiDE, TimeMixer, TimesNet, UMixer}\\
			&\textbf{RevIN-like}: Crossformer, FreTS, LightTS, NLinear, NSformer, SegRNN\\
			&\textbf{Others}: All work related to trends, seasons, hetero-scedasticity, and anomalies below, \\
			\midrule
			\multirowcell{3}{ \small \textbf{Trend} \\ \small \textbf{Seasonality}}&\textbf{Decomposition(Moving Avg)}: Autoformer,  FEDformer, DLinear\\
			&\textbf{Decomposition(Fourier method) and Multi-seasons}: ETSformer, Koopa, TimeMixer\\
			&\textbf{Only Season}: CycleNet, FiLM, Fredformer, FreTS, HI, TimesNet\\
			\midrule
			\multirow{7}{*}{\small \textbf{Volatility}}& \textbf{Decomposition(Moving Avg)}\\
			&\textbf{Fourier method}: FEDformer, FiLM, Fredformer, FreTS, Koopa, TimeMixer, TimesNet \\
			&\textbf{DownSample}: DSformer, MTSMixer, TimeMixer, NHiTS\\
			&\textbf{Data Augmentation}: ETSformer\\
			&\textbf{Representation Dimension:}\\
			&\qquad \textbf{Channel Embedding:} Autoformer, ETSformer, FEDformer, Informer, NSTransformer, TimesNet\\
			&\qquad \textbf{Timestamp Embedding:} All other depth forecasting methods\\
			\midrule
			\multirow{12}{*}{\small \textbf{Memorability}}
			&\textbf{Long short-term: }\\
			& \qquad \textbf{Multi-Scale}: Crossformer, FiLM, Pyraformer, TimeMixer\\
			&\qquad \textbf{Patch}: CATS, Crossformer, LightTS, NHiTS, PatchTST, SegRNN, SparseTSF, Triformer, UMixer\\
			& \qquad \textbf{DownSample}\\
			&\textbf{Channel Strategy: }\\
			& \qquad \textbf{Channel Dependency}: Autoformer, ETSformer, FEDformer, Informer, NSTransformer, TimesNet\\
			&\qquad \makecell[l]{\textbf{Channel Independency}: CycleNet, DLinear, FreTS, HI, Koopa, LightTS, NBeats, NHiTS, NLinear, \\ PatchTST, Pyraformer, SegRNN, SparseTSF, WaveNet}\\
			&\qquad \textbf{Implicit  Channel Interaction}: DeepAR, FiLM, STID, TiDE, Triformer\\
			
			&\qquad \makecell[l]{\textbf{Explicit Channel Interaction}: CATS, Crossformer, DSformer, Fredformer, iTransformer, MTSMixer, \\SOFTS, Sumba, TimeMixer, UMixer}\\
			&\textbf{Transformer-based v.s. MLP-based}\\
			\midrule
			\multirow{4}{*}{\small \textbf{Scedasticity}}& \textbf{Covariate Shift}: ARCH\\
			&\textbf{Residual}: NBeats, NHiTS\\
			&\textbf{Time-(in)variant}: Koopa \\
			&\textbf{RevIN} and \textbf{\textit{RevIN-like}}\\
			\midrule
			\multirow{3}{*}{\small \textbf{Anomaly}}&\textbf{Residual}: NBeats, NHiTS\\
			&\textbf{Time-(in)variant}:Koopa\\
			&\textbf{RevIN}, \textbf{\textit{RevIN-like}}, \textbf{DownSample} and \textbf{Fourier methods}\\
			\hline
			\bottomrule[1.5pt]
		\end{tabular}
	}
\end{table*}

\subsection{Summary of modeling strategies}\label{msb}
We summarize the modeling strategies of existing forecasting models in BasicTS~\cite{shao2023exploring} and align them with time series properties in Table~\ref{ARIES_M}.

\textbf{Stationarity} remains foundational, yet due to limited valid information, most research excluding early financial methods has focused on identifying patterns in non-stationary series. Table~\ref{ARIES_M} offers a simplified categorization, highlighting works on non-stationarity and methods for variance or mean adjustments such as RevIN~\cite{kim2021reversible}.

\textbf{Trend and Seasonality} are frequently coupled through STL decomposition. Moving average methods prioritize trend extraction via sliding window operations before seasonal analysis, while Fourier approaches first identify dominant seasonal components through top-$K$ frequency selection. In addition, some Fourier-based studies neglect trend analysis and focus solely on seasonality.

\textbf{Volatility}-related work first involves a time-series representation perspective, where channel-dependent strategies follow the data representation strategies of computer vision for embedding the channel dimension, and subsequent work often learns temporal information about the timestamp dimension. In addition, methods to alter the inter-temporal volatility such as data smoothing, augmentation, and down-sampling, as well as high-frequency analysis based on Fourier transforms, are also taken into account.

\textbf{Memorability}-related works handle temporal dependencies through attention architectures, multi-scaling, down-sampling, and patch strategies, while the latter three also address short-term dependency through data compression. Another critical aspect is channel strategy: compress channel representations~(dependence), treat channels independently, or emphasize channel interactions. Channel interactions are further categorized into explicit~(e.g., Mixer, iTransformer, traffic forecasting models using cross-channel attention or GNNs) and implicit~(e.g., STID, which assigns channel parameters and relies on regression for self-supervised interaction). Furthermore, the Transformer vs. MLP debate is also related to memorability because of memory capacity.

\textbf{Scedasticity and anomaly} describe the dynamics of numerical transformations in time series. 
Hetero-scedasticity implies covariance shift, high anomalies indicate mean shift, and low anomalies correlate with noise.
Thus, residual decomposition, time-invariant learning and RevIN~(\textit{RevIN-like} ignores variance) are all taken into consideration. Furthermore, noise with smaller anomalies may be closely tied to down-sampling and Fourier methods.

\begin{figure*}[t]
	\centering
	
	\includegraphics[width=0.95\textwidth]{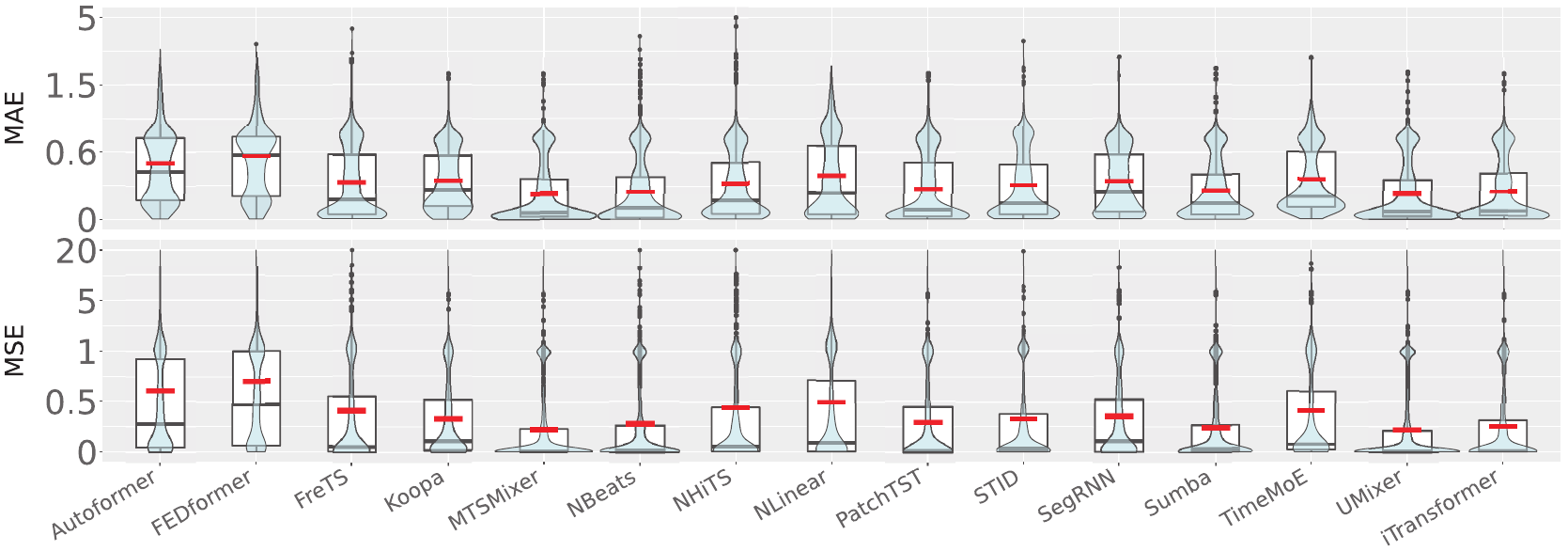}
	\caption{Distribution of MAE and MSE on Synth for several sample models. Different models show distinct boxplot shapes across metrics—for example, Autoformer and MTSMixer—indicating varied data preferences. Meanwhile, similar distributions between models like MTSMixer and UMixer suggest that their shared Mixer backbone leads to similar pattern biases and holds a dominant position over other strategies. In addition, the phenomena in MAE and MSE are unified, and our multiple experiments have proved that such regularity is very stable.}
	\label{maemse}
\end{figure*}

\section{Relation Assessment}\label{relation}
In that section, we attempt to answer: \textit{How do deep modeling strategies favor specific time series properties?} Based on our preparatory work, we present a novel and comprehensive benchmark for deep forecasting models \textit{ARIES TEST}, and assess the relation between modeling strategies and time series properties by analyzing 50+ existing works. 

The follow-up content includes feasibility analysis of relation assessment, baselines, introduction of \textit{ARIES TEST}, relation analysis, and demonstration of strategy analysis based on \textit{ARIES TEST}.

\subsection{Feasibility analysis of relation assessment}\label{fra}
Because strategies have different propensities for properties, the performance distributions between models will diverge and converge.

As shown in Figure~\ref{maemse}, we present the distribution of MAE and MSE of some models in \textit{Synth} via a combination of violin plots and box plots. The violin plots represent the proportion of time series for each metric, with black dots indicating performance outliers of higher proportion. The box plots display the quartiles and medians (bolded in black). Additionally, the mean values of metrics, as always reported in other works, are marked with a red line.

The intuition behind using these plots is that relying solely on mean values of metrics can often be misleading. For instance, Koopa and FreTS exhibit nearly identical mean MAE on \textit{Synth}, yet the median and distribution of metric differ significantly. They may appear to perform similarly based on the mean MAE, but differing distributions suggest that their distinct data preferences, making misapplication on other datasets unlikely to achieve desired results.

Meanwhile, similar metric distributions can be observed between Autoformer and FEDformer, as well as MTSMixer and iTransformer, suggesting that they may have similar data preferences and, in fact, employ shared modeling strategies.

\subsection{Baselines}
To explore the relation between properties and modeling strategies, we select 50+ baselines in BasicTS  \footnote{https://github.com/GestaltCogTeam/BasicTS}~\cite{shao2023exploring}: 

\begin{itemize}
	
	\item\textbf{Traditional local forecasting methods:} AR~\cite{yule1927vii}, MA~\cite{walker1931periodicity}, ARMA~\cite{box1968some}, ARIMA~\cite{hyndman2008forecasting}, ARCH~\cite{akgiray1989conditional}, GARCH, SARIMA, SES, ETS~\cite{chatfield1978holt};
	
	\item\textbf{Machine learning methods:} SVR, PolySVR~\cite{boser1992training}, CatBoost~\cite{prokhorenkova2018catboost}, LightGBM~\cite{ke2017lightgbm};
	
	\item\textbf{Transformer-based deep learning methods:}
	Autoformer~\cite{wu2021autoformer}, Crossformer~\cite{zhang2023crossformer}, DSFormer~\cite{yu2023dsformer}, ETSformer~\cite{woo2022etsformer}, FEDformer~\cite{zhou2022fedformer}, Fredformer~\cite{piao2024fredformer}, Informer~\cite{zhou2021informer}, iTransformer~\cite{liuitransformer}, NSformer~\cite{liu2022non}, PatchTST~\cite{nietime}, Pyraformer~ \cite{liu2021pyraformer}, Triformer~\cite{cirsteatriformer};
	
	\item\textbf{MLP-based deep learning methods:} CATS\cite{DBLP:conf/nips/Kim00K24}, CycleNet~\cite{lin2024cyclenet}, DLinear~\cite{zeng2023transformers}, FiLM~\cite{zhou2022film}, FreTS~\cite{yi2024frequency}, Koopa~\cite{liu2022non}, LightTS~ \cite{zhang2022less}, MTSMixer~\cite{li2023mts}, NBeats~\cite{oreshkinn}, NHiTS~\cite{challu2023nhits}, NLinear~\cite{zeng2023transformers}, SOFTS~\cite{DBLP:conf/nips/LuCYZ24}, SparseTSF~\cite{linsparsetsf}, STID~\cite{shao2022spatial}, TiDE~\cite{daslong}, TimeMixer~\cite{wangtimemixer}, TimesNet~\cite{wutimesnet}, UMixer~\cite{ma2024u};
	
	\item\textbf{Foundational models:} MOIRAI~(Base \& Large)~\cite{woo2024moirai}, Time-MoE~\cite{shi2025timemoe}
	
	\item\textbf{Others:} DeepAR~\cite{salinas2020deepar}, HI~\cite{cui2021historical}, SegRNN~\cite{lin2023segrnn}, Sumba~\cite{chen2024structured}, WaveNet~\cite{van2016wavenet};
	\vspace{-10pt}
\end{itemize}

\begin{figure*}[t]
	\centering
	\includegraphics[width=0.95\textwidth]{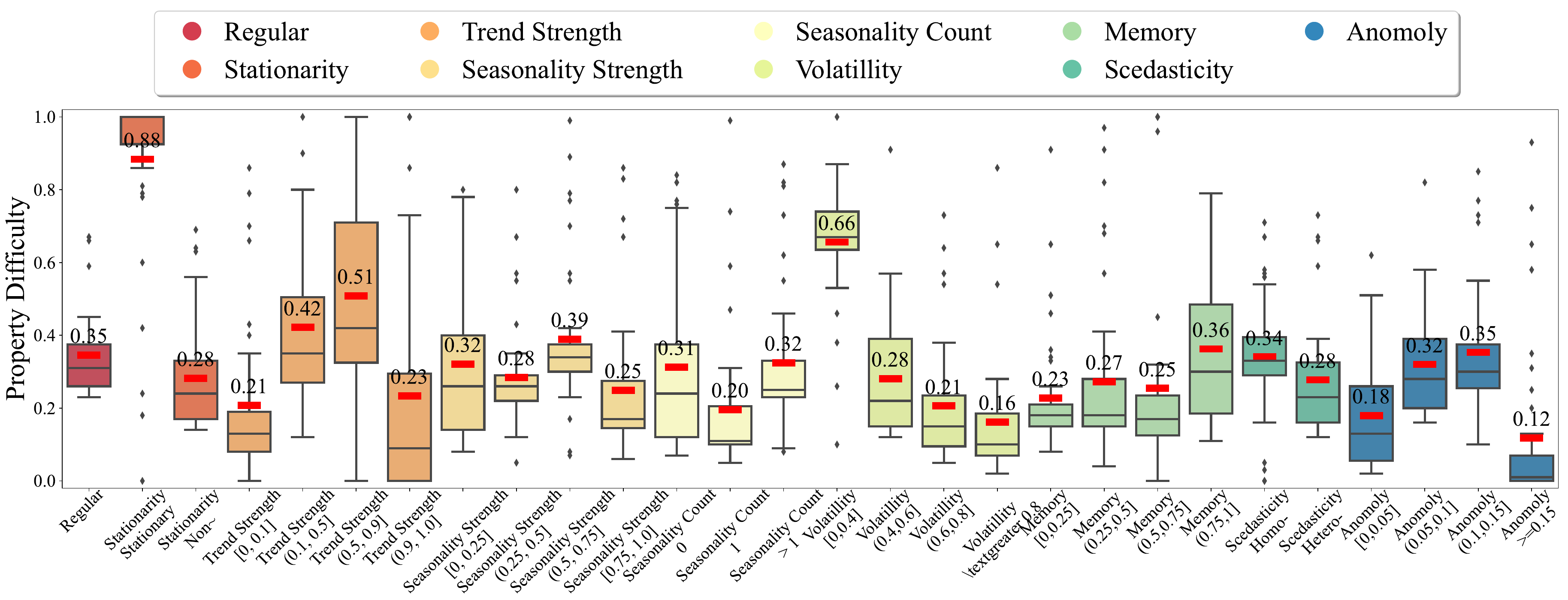}
	\caption{Results of the difficulty of fine-grained time series properties across forecasting models. Encouragingly, the notion of difficulty exists with respect to properties and models. Patterns such as strong trend and strong memory show disagreement across models, directly indicating a profound connection between data properties, and modeling strategies. ARIES aims to investigate these connections, which may further inspire future research in time series forecasting.}
	\label{difficulty}
\end{figure*}

\subsection{Introduction of \textit{ARIES TEST}}

\textbf{Dataset:} \textit{Synth} is divided into training~(70\%), validation~(10\%), and test~(20\%) sets.
The length of historical and future series is 336, and the step size of the data sliding window is 1. The history of each series of test set will be computed by evaluating time series properties in Section~\ref{etsp}.

\textbf{Property cleaning:} We re-emphasize that we exclude stationary series when discussing other properties, including in Figure~\ref{datasetsbing}, as they typically lack learnable information. This has been supported by prior financial studies~\cite{box1968some} and empirically validated in subsequent experiments.

\textbf{Performance Reporting: }Each baseline model is tuned to its optimal hyper-parameters, with multiple runs performed to select the model parameters corresponding to the median performance. The forecasting performance on each test-set series is then recorded using these parameters.
To more intuitively reflect the performances of all models, we report the \textit{Mean} and \textit{Median} of \textit{MAE} and \textit{MSE} in Table~\ref{tab1} and Table~\ref{tab2}, and highlight the top 10 entries in blue and the bottom 10 in red  with darker shades highlighting leads/lags.

\textbf{Goal:} \textit{ARIES TEST} validates and visualizes the performance of each model under diverse properties in \textit{Synth} for more in-depth forecasting modeling studies.

\textbf{Tutorial:} If researchers want to evaluate a new model’s performance and pattern preferences, they only need to submit test-set performance logs under BasicTS’s identical configuration. ARIES TEST will then return the corresponding results from Tables~\ref{tab1} and ~\ref{tab2}.

\subsection{Relation Analysis}\label{ana}

\textbf{Overview of property difficulty:} With the model performance results from Tables~\ref{tab1} and \ref{tab2}, we present Figure~\ref{difficulty} to illustrate the difficulty of each fine-grained property, based on the min-max normalized ratio of each model’s Mean MAE to the $Regular$. The boxplot displays the quartile distributions of difficulty with whiskers denoting normal ranges and diamond markers denoting statistical outliers. Properties of the same category are color-coded, with red lines marking the mean difficulty among baselines.

Notably, wide interquartile ranges~(IQR) or whisker spans signify substantial model performance variance such as strong trends and long-term memory, \textbf{revealing distinct modeling strategy preferences for specific time series properties.}

In the subsequent sections, we will assess the relation between each property and modeling strategies, include the overview of model performance for that property, the analysis in conjunction with the specific modeling strategy, and the key findings that reveal problems and possible innovations.

\begin{table*}[ht]	 		 		
	\centering 		 		 		
	\caption{Part 1 of the results between the model performance and the time series properties in ARIES TEST}
	\renewcommand\arraystretch{0.6}
	\tiny
	\setlength{\tabcolsep}{1.8pt}
	\label{tab1}
	\scalebox{0.92}
	{
\begin{tabular}{|>{\centering}p{37pt}|>{\centering}p{12pt}|>{\centering}p{14pt}>{\centering}p{14pt}|>{\centering}p{14pt}>{\centering}p{14pt}|>{\centering}p{14pt}p{14pt}  |>{\centering}p{14pt}p{14pt}|>{\centering}p{14pt}p{14pt}|>{\centering}p{14pt}p{14pt}|>{\centering}p{14pt}p{14pt}|>{\centering}p{14pt}p{14pt}|>{\centering}p{14pt}p{14pt}|>{\centering}p{14pt}p{14pt}|>{\centering}p{14pt}p{14pt}|>{\centering}p{14pt}p{14pt}|>{\centering}p{14pt}p{14pt}|>{\centering}p{14pt}p{14pt}|
			}
			\hline
			\multicolumn{2}{|c|}{\textbf{Model}} & \multicolumn{2}{c|}{\textbf{Regular}}&\multicolumn{4}{c|}{\textbf{Stationarity}}&\multicolumn{8}{c|}{\textbf{Trend Strength}}&\multicolumn{8}{c}{\textbf{Seasonality Strength}}&\multicolumn{6}{|c|}{\textbf{Seasonality Count}}\\
			
			\multicolumn{2}{|c|}{\textbf{}}&&&\multicolumn{2}{c}{\textbf{Stationary}}&\multicolumn{2}{c|}{\textbf{Non~}}&\multicolumn{2}{c}{\textbf{[0, 0.1]}}&\multicolumn{2}{c}{\textbf{(0.1, 0.5]}}&\multicolumn{2}{c}{\textbf{(0.5, 0.9]}}&\multicolumn{2}{c|}{\textbf{(0.9, 1.0]}}&\multicolumn{2}{c}{\textbf{[0, 0.25]}}&\multicolumn{2}{c}{\textbf{(0.25, 0.5]}}&\multicolumn{2}{c}{\textbf{(0.5, 0.75]}}&\multicolumn{2}{c}{\textbf{(0.75, 1.0]}}&\multicolumn{2}{|c}{\textbf{0}}&\multicolumn{2}{c}{\textbf{1}}&\multicolumn{2}{c|}{\textbf{\(\geq\) 1}}\\
			\hline
			\multicolumn{2}{|c|}{\textbf{}}&Mean&Med.&Mean&Med.&Mean&Med.&Mean&Med.&Mean&Med.&Mean&Med.&Mean&Med.&Mean&Med.&Mean&Med.&Mean&Med.&Mean&Med.&Mean&Med.&Mean&Med. &Mean&Med.\\ 
			\hline
			\multirow{2}{*}{ \textbf{ARCH}}& MAE&0.79&0.804&0.774&0.794&0.792&0.807&0.852&0.829&0.883&0.835&1.013&0.863&0.519&0.249&0.682&0.413&0.732&0.794&0.823&0.823&0.861&0.834&0.655&0.357&0.87&0.833&0.843&0.827\\
			&MSE&2.378&1.0&1.034&0.986&2.537&1.001&1.204&1.05&1.288&1.055&3.782&1.083&6.265&0.068&5.083&0.19&0.929&0.99&1.156&1.045&1.217&1.055&5.082&0.14&1.603&1.013&1.2&1.064\\
\hline
	\end{tabular}}
\end{table*}

\begin{table*}[t]	 		 		
	\centering 		 		 		
	\caption{Part 2 of the results between the model performance and the time series properties in ARIES TEST.}
	\renewcommand\arraystretch{0.6}
	\tiny
	\setlength{\tabcolsep}{1.8pt}
	\label{tab2} 
	\scalebox{0.92}{	 	 			 		
		\begin{tabular}{|>{\centering}p{37pt}|>{\centering}p{12pt}|>{\centering}p{14pt}>{\centering}p{14pt}|>{\centering}p{14pt}>{\centering}p{14pt}|>{\centering}p{14pt}p{14pt}  |>{\centering}p{14pt}p{14pt}|>{\centering}p{14pt}p{14pt}|>{\centering}p{14pt}p{14pt}|>{\centering}p{14pt}p{14pt}|>{\centering}p{14pt}p{14pt}|>{\centering}p{13pt}p{13pt}|>{\centering}p{13pt}p{13pt}|>{\centering}p{14pt}p{14pt}|>{\centering}p{14pt}p{14pt}|>{\centering}p{14pt}p{14pt}|>{\centering}p{14pt}p{14pt}|
			}
			\hline
			\multicolumn{2}{|c|}{\textbf{Model}} & \multicolumn{8}{c|}{\textbf{Volatility}}&\multicolumn{8}{c|}{\textbf{Memory}}&\multicolumn{4}{c|}{\textbf{Scedasticity}}&\multicolumn{8}{c|}{\textbf{Anomaly}}\\
			
			\multicolumn{2}{|c|}{\textbf{}}&\multicolumn{2}{c}{\textbf{[0,0.4]}}&\multicolumn{2}{c}{\textbf{(0.4,0.6]}}&\multicolumn{2}{c}{\textbf{(0.6,0.8]}}&\multicolumn{2}{c|}{\textbf{\textgreater 0.8}}&\multicolumn{2}{c}{\textbf{[0,0.25]}}&\multicolumn{2}{c}{\textbf{(0.25,0.5]}}&\multicolumn{2}{c}{\textbf{(0.5,0.75]}}&\multicolumn{2}{c|}{\textbf{(0.75,1]}}&\multicolumn{2}{c}{\textbf{Homo-}}&\multicolumn{2}{c|}{\textbf{Hetero-}}&\multicolumn{2}{c}{\textbf{[0,0.05]}}&\multicolumn{2}{c|}{\textbf{(0.05,0.1]}}&\multicolumn{2}{c|}{\textbf{(0.1,0.15]}}&\multicolumn{2}{c|}{\textbf{\textgreater 0.15}}\\
			\hline
			\multicolumn{2}{|c|}{\textbf{}}&Mean&Med.&Mean&Med.&Mean&Med.&Mean&Med.&Mean&Med.&Mean&Med.&Mean&Med.&Mean&Med.&Mean&Med.&Mean&Med.&Mean&Med.&Mean&Med.&Mean&Med.&Mean&Med. \\ 
			\hline
			\multirow{2}{*}{ \textbf{ARCH}}& MAE&0.837&0.809&0.712&0.787&0.852&0.837&1.037&0.875&0.79&0.817&0.822&0.821&0.877&0.855&0.758&0.618&0.466&0.238&0.818&0.815&0.778&0.82&0.774&0.801&0.83&0.812&0.832&0.788\\
			&MSE&1.227&1.012&0.961&0.972&1.691&1.026&15.201&1.226&1.052&0.999&1.162&1.068&1.309&1.187&5.392&0.488&0.462&0.061&2.698&1.009&1.502&0.898&4.256&0.981&1.243&1.061&1.146&1.008\\
\hline
			
		\end{tabular}
	}
	
\end{table*}

\vspace{10pt}
\subsubsection{\textbf{Stationarity}}\

While stationarity theoretically implies time-invariant statistical properties, the results in Figure~\ref{difficulty} directly show that stationarity is instead \textbf{the most difficult} due to lack of learnable information.

\textbf{Strategy Analysis}: As evidenced by Figure~\ref{difficulty} and Table~\ref{tab1}, all models perform poorly on stationary series. 
NBeats demonstrates marginal superiority with merely 4.5\% mean improvement over suboptimal performers and no significant median enhancement, underscoring the intrinsic unpredictability of stationary series.

\textbf{Key Findings}: Both ARIES TEST and empirical evidence conclusively demonstrate that stationary time series with low signal-to-noise ratios (SNR) lack learnability due to the absence of discernible temporal patterns (e.g., trends, seasonality, or memorability).

\vspace{10pt}
\subsubsection{\textbf{Trend}}\

\textbf{Easy to learn:} No trend~([0, 0.1]); 
\textbf{Relative Difficulty: } Moderate trends~((0.1, 0.9]) due to potential seasonal coupling;
\textbf{Divergence: } Trends exist~((0.1, 1]), especially strong ones~((0.9, 1])

\textbf{Strategy Analysis}: Regarding the decomposition strategies in Table~\ref{ARIES_M}, when comparing performance on $Trend$ $Strength$ to $Regular$ performance, moving average approaches slightly outperform Fourier methods. However, neither approach demonstrates clear advantages over less trend-correlated strategies like patching or down-sampling.

Moreover, most MLP-based models fail to handle strong trends, and the addition of RevIN~\cite{kim2021reversible} exists to enhance this remarkable property. Observing the work that is near-perfect at forecasting strongly trending sequences, the ultimate solution to the topic appears to be channel interaction strategies such as the Mixer architecture, which may be attributed to enhanced long-term dependency modeling.

\textbf{Key Findings}: Moving  average approach of decoupling strategies favors strong trends over the Fourier-based ones, but the effectiveness of decomposition is questionable and needs to be explored in depth; Models relying only on MLPs cannot handle them, but the most suitable strategies are in fact those related to channel interactions.
\vspace{10pt}
\subsubsection{\textbf{Seasonality}}\

\textbf{Easy to learn:} Strong~((0.75, 1]) seasons, or single season;
\textbf{Relative Difficulty: } Moderate~((0.25, 0.75]) or multiple seasons;
\textbf{Divergence: } Weak~((0, 0.25]) or no seasonality because they are just strong trends;

\textbf{Strategy Analysis}:  While Frequency-domain methods like the Fourier transform have historically dominated seasonal modeling, early implementations demonstrated limited efficacy. Fortunately, recent models such as FiLM~\cite{zhou2022film}, FreTS~\cite{yi2024frequency} and Fredformer~\cite{piao2024fredformer} achieve superior performance by mitigating high-frequency noise interference~\cite{zhou2022film} and frequency bias~\cite{piao2024fredformer, ye2024frequency}. 

Furthermore, the residual mechanism of NBeats have been witnessed to be effective in capturing seasonality. We believe this is due to the fact that the deep residuals of the black box effectively separate the learnable seasonal components, whereas explicit decoupling methods have not been able to achieve the same effect.

\textbf{Key Findings}: Fourier-based methods are effective in extracting seasonal information, but need to be mindful of frequency domain issues such as noise. Comparing the failure of MLP methods on trends, they did not lead Transformer-based methods on seasonality. Deep residuals are currently the leading strategy although only adopted by NBeats, and are a technique worth exploring in the future.

\vspace{10pt}
\subsubsection{\textbf{Volatility}}\

\textbf{Easy to learn:} High~(\textgreater 0.8) volatility;
\textbf{Relative Difficulty: } Low~((0, 0.4])  volatility;
\textbf{Divergence: } Moderate~((0.4, 0.8]) volatility;

\textbf{Strategy Analysis}:
Counterintuitively, deep learning models perform better with increasing volatility, as evidenced in Figure~\ref{difficulty} and Table~\ref{tab2}, but this phenomenon is not observed in local forecasting methods sharing identical pre-processing pipelines. We conjecture that discernible pattern variation in low volatility data is limited.

Moreover, the representational dimensions of time series in deep forecasting methods—particularly timestamps and channel values—require careful consideration. Early forecasting approaches adopted channel-dependent strategies by embedding channel dimensions through computer vision-inspired correlations, but this inadvertently neglected temporal pattern learning. Empirical performance analysis reveals that such channel-dependent methods exhibit no clear preference under increasing volatility regimes, whereas deep forecasting methods that explicitly learn timestamp representations demonstrate superior capability in handling high-volatility scenarios.

The other potentially relevant strategies such as moving average, downsampling, and data augmentation in Table~\ref{ARIES_M} have limited enhancement. While FreTS, and Fredformer with Fourier methods showing the most significant benefits in high-volatility. Moreover, patch and channel interaction strategies also perform well through enhanced pattern integration.

\textbf{Key Findings}: The preference for high-volatility scenarios constitutes a key advantage of deep forecasting over traditional methods. Sole reliance on channel-dimensional representations proves insufficient, necessitating explicit timestamp embedding. While frequency-domain approaches, patch-based strategies, and channel interactions demonstrate clear high-volatility adaptability, they require further systematic investigation.

\vspace{10pt}
\subsubsection{\textbf{Memory}}\

\textbf{Easy to learn:} Low and moderate~((0, 0.75]) memory;
\textbf{Relative Difficulty \& Divergence: } High~((0.75, 1]) memory;

Experimental results in Table~\ref{tab2} and Figure~\ref{difficulty} reveal that as memory increases, which means series exhibit stronger long-term dependence, the performance of most models declines more significantly than with any other property. However, parameter-free HI maintains robust performance, \textbf{implying that memorability is a very important and unique topic in deep time series forecasting}.

\vspace{10pt}
\textbf{Long short-term}: Multi-scale, down-sampling, and patching strategies excel at short-term dependencies~([0, 0.75]) but struggle with long-term patterns, except SegRNN and Mixer variants. Moreover, models like Crossformer and Triformer, designed for long-term dependencies, exhibit performance decay on such series. In contrast, ETSformer and NSTransformer demonstrate enhanced long-term capability without these strategies compared to their $Regular$ performance.

\vspace{10pt}
\textbf{Channel Strategy}: 
Early discussions on channel dependence versus independence primarily revolved around the trade-off between effectively utilizing channel-specific information and the risk of overfitting. Our experimental findings align with the conclusions drawn in PatchTST, reinforcing the superiority of channel-independent approaches~\cite{DBLP:journals/tkde/HanYZ24, DBLP:conf/iclr/ZhaoS24}, even on larger datasets like \textit{Synth}.

Subsequent studies have refocused on channel interactions, particularly how this information is learned. Implicit methods, such as STID and Triformer that use learnable channel embeddings, perform well on short-term dependencies but limit long-term memorization. In contrast, explicit interaction methods, such as DSformer, Fredformer, iTransformer, Mixer variants and so on, leverage attention and GNNs to model channel relationships, demonstrating superior performance on long-term dependencies.

\vspace{10pt}
\textbf{Transformer vs. MLP: }NLinear and DLinear outperform early Transformer-based models, reigniting debates on the utility of Transformers for time series forecasting. However, our experiments on memorability reveal that most MLP-based models struggle with long-term memory, except for Mixer variants that incorporate channel interactions, while the attention mechanism of Transformer is well-suited for modeling long-term dependencies.

\vspace{10pt}
\textbf{Key Finding: }Strategies for short-term dependence enhancement may slightly compromise long-term dependence modeling. ARIES attributes both structure and channel strategy to the memory capacity. Channel dependence or Transfomer tends to face overfitting risks  due to more parameters, while channel independence or MLP-based models often exhibit weaker fitting abilities.
However, channel strategy is more important than structure, and there is now minimal performance gap between Transformer and MLPs that adopt novel channel interactions.

\vspace{10pt}
\subsubsection{\textbf{Scedasticity}}\

Traditional methods and early depth forecasting, consider homo-scedasticity to be simpler, but this conclusion is reversed when depth strategies for mitigating hetero-scedasticity are proposed.

\textbf{Strategy Analysis}: Parameter-free RevIN has emerged as a fundamental component for handling hetero-scedasticity. As Table~\ref{tab2} evidences, RevIN-enhanced models consistently outperform legacy architectures on hetero-scedastic data. However, RevIN-like variants focusing solely on mean adjustment remain ineffective against covariance shift due to variance neglect.

Koopa, NBeats and NHiTS employ distinct strategies to address shift, positioning them at opposite ends of the performance spectrum. Koopa handles time-variant and time-invariant information separately, achieving outstanding performance on homo-scedastic data but facing challenges in extracting meaningful information from hetero-scedasticity. In contrast, NBeats and NHiTS, with their temporal residual learning, excel at capturing hetero-scedastic patterns but struggle with homo-scedasticity due to pattern confusion caused by excessive differencing.

\textbf{Key Finding: } Mitigating distributional shift and modeling variability are important for alleviating hetero-scedasticity, and they should become key focuses for future research.

\vspace{10pt}
\subsubsection{\textbf{Anomaly}}\

\textbf{Easy to learn:} High anomaly~(\textgreater 0.15);
\textbf{Relative Difficulty: } Moderate anomaly~((0.05, 0.15]);
\textbf{Divergence: } Low anomaly~([0, 0.05]);

\textbf{Strategy Analysis}: 
NBeats achieves SOTA performance on series with high anomalies, but its performance on low-intensity anomalies experiences a relative decline, as its hierarchical feature extraction struggles to provide meaningful benefits in stable sequences. Additionally, novel down-sampling and Fourier-based methods such as DSformer and Fredformer, as well as Koopa, show no extra improvements on data with high anomalies.

For the high-anomaly distribution shift issue, RevIN indeed brings significant improvements. However, considering the still prevalent lower anomalies with noise, RevIN is not applicable , and we find that MLP-based methods even with techniques like Patch and frequency-domain analysis still fail to effectively model these interference-laden scenarios.

\textbf{Key Finding: }RevIN and RevIN-like methods with decoupled means are effective for high anomalies. Techniques like decoupled variability and downsampling do not provide significant advantages for this property, and MLP-only architectures are not recommended for lower anomalies.

\subsubsection{Extra focus on foundation models}
Given the absence of clear strategy and resource constraints, time series foundation models are not currently a priority for ARIES. However, zero-shot testing of Moriai and TimeMoE still revealed interesting findings.

Moirai's performance only approaches that of the parameter-free HI, while TimeMoE's performance has reached a level between TimeNet and FreTS.

Moirai does not demonstrate property preferences similar to trained deep forecasting models. For instance, it shows no particular inclination towards strong seasonality or volatility patterns, and also finds homo-scedasticity more challenging, which reminiscent of traditional methods. TimeMoE exhibits preferences similar to MLP-based models, yet likewise displays no volatility bias, which may be related to its foundational model's encoding mechanism.

Notably, in a rather counterintuitive observation, the smaller-parameter Moirai-base handles strong-trend sequences effectively, while both Moirai-large and TimeMoE show significant performance degradation. This likely occurs because the more complex foundational models tend to overcomplicate simple patterns, leading to erroneous handling.

\begin{table*}[t]
	\centering 		 		 		
	\caption{Modified relation on modeling strategy and time series properties.}
	\renewcommand\arraystretch{1}
	\label{ARIES_M2} 	
	\resizebox{1\linewidth}{!}{
		\begin{tabular}{c|c|l}
			\toprule[1.5pt]
			\hline
			\small \textbf{Data Property}& \textbf{Granularity} &\small \textbf{Modeling Strategy} \\
			\midrule
			\hline
			\multirow{2}{*}{\small \textbf{Stationarity}}&\textbf{Stationary}& Unlearnable\\ \cline{2-3}
			&\textbf{Non-stationary}& All Deep forecasting Models\\ 
			\midrule
			
			\multirowcell{5}{ \small \textbf{Trend} \\ \small \textbf{Seasonality}}&\textbf{General }&\textbf{RevIN}, \textbf{\textit{RevIN-like}}, \textbf{Channel Interaction}\\ \cline{2-3}
			&\multirowcell{2}{\textbf{Strong Trend}}&\textbf{Decomposition(Fourier method), Transformer backbone}\\ 
			&&\textit{Avoid: } Decomposition(Moving Avg), MLP-only backbone, Complex Foundation Model\\ \cline{2-3}
			&\multirowcell{2}{\textbf{Strong Seasonality \& Multi-season}}&\textbf{Decomposition(Moving Avg)}, \textbf{Only Season, Residual}\\
			&& \textit{Avoid: } Decomposition(Fourier method)\\
			\midrule
			\multirow{3}{*}{\small \textbf{Volatility}}& \textbf{Low Volatility}& Difficult to learn\\ \cline{2-3}
			&\multirowcell{2}{\textbf{High Volatility}}&\textbf{Timestamp Embedding}, \textbf{Fourier method, RevIN} \\
			&&\textit{Avoid}: Channel Embedding, Time Series Foundation Model\\
			\midrule
			\multirow{5}{*}{\small \textbf{Memorability}}
			&\multirowcell{2}{\textbf{General}}&\textbf{RevIN}, \textbf{\textit{RevIN-like}}, \textbf{Channel Indepency, Channel Interaction}\\ 
			&& \textit{Avoid:} Channel Dependency\\ \cline{2-3}
			&\textbf{Short-term dependence}&\textbf{DownSample, Multi-Scale, Patch, MLP backbone}\\ \cline{2-3}
			&\multirowcell{2}{\textbf{Long-term dependence}}& \textbf{Transformer backbone}\\
			&&\textit{Avoid: } MLP-only backbone\\
			\midrule
			\multirow{3}{*}{\small \textbf{Scedasticity}}&\multirowcell{2}{\textbf{Homo-Scedasticity}} &\textbf{Time-(in)variant}\\
			&&\textit{Avoid: } Residual\\ \cline{2-3}
			&\multirowcell{2}{\textbf{Hetro-Scedasticity}}&\textbf{RevIN, Residual}\\
			&&\textit{Avoid: } Time-(in)variant\\
			\midrule
			\multirow{4}{*}{\small \textbf{Anomaly}}&\multirowcell{2}{\textbf{Low anomaly}}& \textbf{DownSample, Fourier method, Patch}\\ 
			&&\textit{Avoid: } MLP-only backbone\\ \cline{2-3}
			&\textbf{High anomaly}&\textbf{RevIN}, \textbf{\textit{RevIN-like}}, \textbf{Residual}\\
			\hline
			\bottomrule[1.5pt]
		\end{tabular}
	}
\end{table*}

\subsection{Summary of Relation Assessment}
Based on our previous analysis of the relations between each property and modeling strategies, we have revised Table~\ref{ARIES_M} and created a simplified Table~\ref{ARIES_M2} to facilitate quick reference and understanding of key findings. Due to the limited number of models and incomplete strategy ablation, the results in Table~\ref{ARIES_M2} still need to be examined in depth by follow-up work.

Tentatively, we use ARIES TEST to analyze two common modeling strategies: seasonal-trend decomposition and RevIN. ARIES identifies their associated time series properties and, for the first time, reveals that different decomposition strategies exhibit contrasting property preferences. Due to space limitations, detailed discussions are provided in Appendix C.

\section{Model Recommendation}\label{model}

In specialized domains, time series forecasting still predominantly relies on traditional methods, primarily because mathematical-statistical approaches typically include comprehensive data analysis reports and model parameter interpretations. These elements serve as crucial psychological reassurance during human decision-making processes. Consequently, interpretable recommendations are crucial for forecasting models, representing a significant bottleneck for the widespread adoption of deep forecasting methods.

This section presents the first recommendation framework for deep forecasting models proposed by ARIES, including feasibility analysis, model recommendation algorithms, sample experiments, and recommendation results analysis.

\subsection{Feasibility analysis of model recommendation}\label{fmr}
iTransformer is recognized as a state-of-the-art~(SOTA) model for time series forecasting. While it performs near-optimally on every metric, it falls short on certain specific patterns. Based on \textit{Synth}, we select time series with no trend ([0, 0.1]), low volatility ([0, 0.6]), and low to medium anomalies ([0, 0.15]) to compare iTransformer with NLinear, a simple MLP-based model, across varying memorability.

\begin{figure}[htb]
	\centering
	\includegraphics[width=0.43\textwidth]{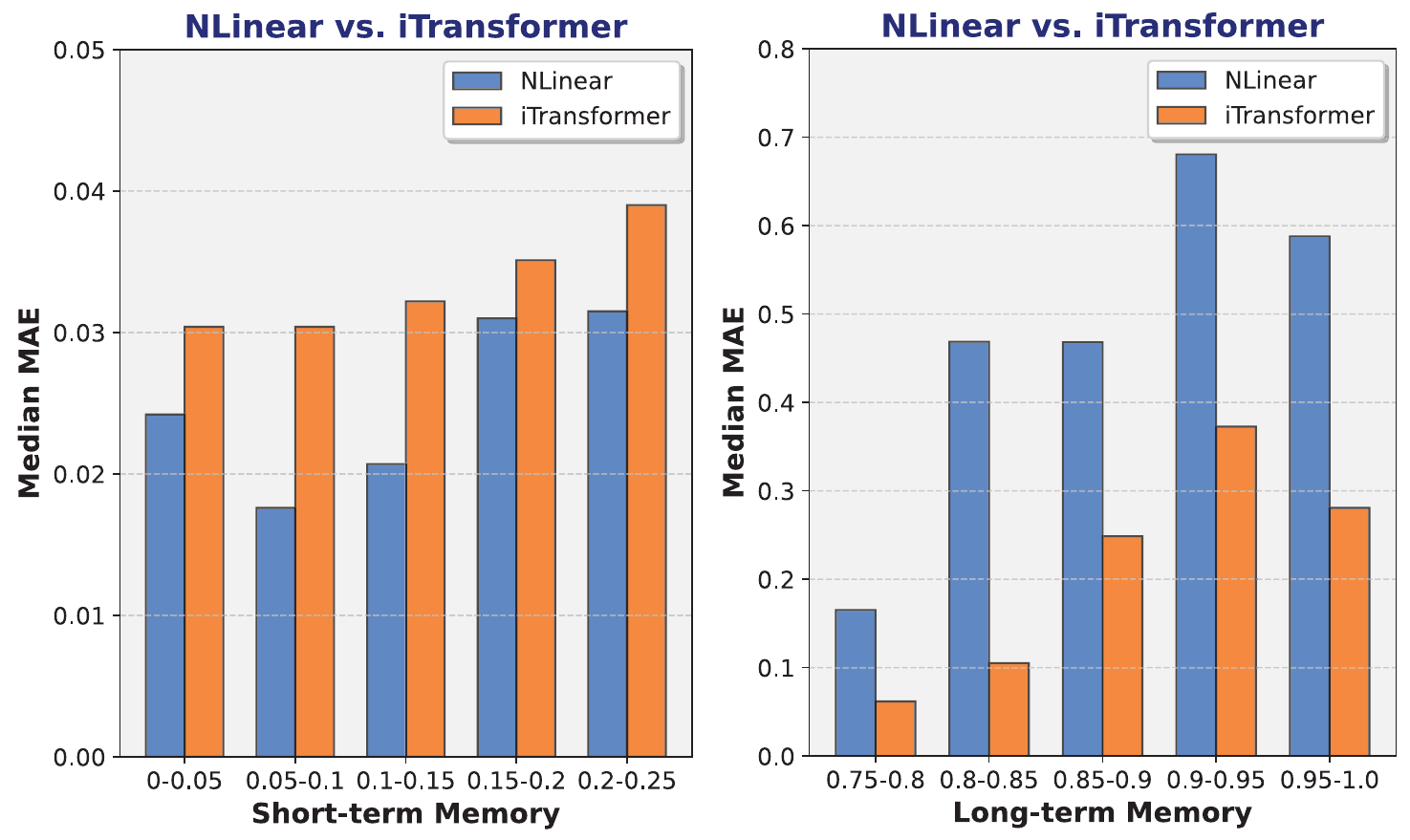}
	\caption{Performance difference between NLinear and iTransformer on different memorization on low trend strength, low anomaly series.}
	\label{recommend2}
\end{figure}

Figure~\ref{recommend2} reports their median MAE on weak and strong memorability. Although iTransformer excels in individual metrics like no trend, low volatility, and low anomalies, NLinear outperforms it on series with low memorability (short-term dependency).
Conversely, iTransformer dominates when series exhibit long dependency. This highlights that models have inherent biases toward specific data patterns. However, real-world time series often involve complex combinations of simple properties, underscoring the critical role of model recommendation in practical time series forecasting.

\subsection{Recommendation Algorithm}
In this section, we introduce an information retrieval-based recommendation algorithm for deep forecasting models. Briefly, it consists of: (1) a key-value storage mapping \textit{Synth} to forecasting performance, (2) query construction using real-world data, and (3) a retrieval and ranking-based recommendation method. The pseudo-code is shown in Algorithm 2 of Appendix D, which resembles a pure collaborative filtering approach~\cite{yang2024discrete}. Moreover, we will provide interpretable rationales for the recommendation results.

\vspace{10pt}
\subsubsection{Synthetic Data-Model Performance Mapping}\label{sdmpm} 
\ 

\textbf{Initial Construction of Keys}: The historical segment $\mathcal{S}$ of \textit{Synth}'s test set serves as the $Key$ for recommendation. Given \textit{Synth}'s dimensions $B\times N\times L$~(batch size $\times$ series count $\times$ series length), the total valid keys are $B\times N$.

\textbf{Value Construction}: For each series, model performance $\mathcal{M} = (Model, MAE, MSE)$ are stored as $Value$, forming key-value pairs $\mathcal{KV}$: \textit{Synth}$_{i, j}$ $=$ $\{(Model_{1}, MAE_{1 ,i, j}, MSE_{1  i, j}),...\}$, where $i, j$ are batch and count.

\textbf{Key Compression}: To enable efficient practical dataset alignment, each \textit{Synth} series is compressed into an 8-bit property vector $\mathbf{p}$ via interval binning strategy $\mathcal{B}$ from Figure~\ref{datasetsbing} to replace keys in $\mathcal{KV}$. For instance, a memorability score of 0.33 maps to the second quartile.

\vspace{10pt}
\subsubsection{Evaluation of the Properties of Practical Time Series}\label{eppts}
\ 

Given practical time series dataset $\mathcal{Q}$ with dimensions $B^{'} \times N^{'} \times L^{'}$, we process the historical segments for property evaluation. While $B^{'}$ and $N^{'}$ can be arbitrary, we suggest that the length $L^{'}$ is sufficiently large to ensure reliable property analysis and model recommendation.

\textbf{Query Construction}: Following the \textbf{Key Compression} protocol, series in $\mathcal{Q}$ are transformed into the same 8-bit property vectors $\mathbf{p}'$, replacing raw temporal data for similarity search.

\vspace{10pt}
\subsubsection{Similarity Search and Performance Ranking}\

Our recommendation method operates through the constructed representations of practical data (\textbf{Query}), synthetic data \textit{Synth} (\textbf{Key}) and corresponding model performances (\textbf{Value}).

\textbf{Similarity Search}: Implements a three-stage process:

\begin{itemize}
	\item \textbf{Query Grouping}: Aggregate the number of identical queries $\mathbf{p}'$ in $\mathcal{Q}$ as the weight for following vector retrieval and performance sampling to save computational resources.
	
	\item \textbf{Vector Retrieval}: Query $\mathbf{p}'$ for the most similar $Keys$ in  $\mathcal{KV}$ with the nearest neighbors method~\cite{kolbe2010efficient}, which permits altering non-significant patterns such as moderate memorability or seasonality to expand the search for neighbors.
	\item \textbf{Performance Sampling}: Retrieve corresponding model performances $\tilde{V}$ with the given sampling rate $\tau$, allowing repeats to ensure that each query $\mathbf{p}'$ is treated fairly.
\end{itemize}

\textbf{Performance Ranking}: Entails a two-step procedure:
\begin{itemize}
	\item \textbf{Individual Recording}: Record the sampled performance of all models for each selected $\tilde{v}'$. 
	\item \textbf{Overall Ranking}: Compute and rank the mean of performance across all $\mathbf{p}'$ to recommend appropriate forecasting models for the practical time series dataset $\mathcal{Q}$. 
\end{itemize}

\vspace{10pt}
\subsubsection{Interpretability suggestions for properties, strategies and models}\

We analyze intermediate data from the recommendation method to provide convincing explanations for the recommendation results, these interpretable analyses will be presented to the user as additional recommendations in Table~\ref{recom}:
\begin{itemize}
	\item \textbf{Property:} Based on property evaluation of practical dataset in Query Construction, we provide the most remarkable properties of the data and their percentages to explicitly inform users of the prominent patterns. 
	\item \textbf{Strategy:} According to the prominent patterns, ARIES recommends the preferred strategies for the practical dataset based on the property-strategy relationships in Table~\ref{ARIES_M2}, prioritizing them in order.
	\item \textbf{Model:} By comparing the improvement and degradation of query performance in Overall Ranking with $Regular$ results in Table~\ref{tab1}, ARIES can identify models with potential preferences and models to avoid.
\end{itemize}

In addition to model recommendations, ARIES provides supplementary model suggestions: Given the narrow performance gaps among state-of-the-art deep forecasting models, while fully relying on forecasting performance for model retrieval may align with real-world needs, we advocate granting early-stage models more opportunities. This is because hyperparameter tuning and incorporating simple yet advanced strategies could significantly enhance their performance.

\subsection{Experimental Setup}
\textbf{Datasets: }We validate recommendation efficacy on three benchmark datasets Electricity, ETTh1, and PEMS08 under realistic constraints. We partition the datasets temporally into training, validation, and test sets with ratios of 70\%/10\%/20\% for Electricity and ETTh1, and 60\%/20\%/20\% for PEMS08, using fixed-length 336-step windows. To prevent data leakage, we only analyze test set historical segments for property evaluation and model recommendation.

\textbf{Forecasting Models:} Given the training inefficiency of traditional methods, we focus exclusively on deep learning models. For all forecasting models, we adopt the suggested parameters from BasicTS. Considering that the hyperparameters in this benchmark may not be optimal, this could lead to ARIES' fine-ranking recommendations deviating slightly from real-world expectations due to subtle performance gaps.

\textbf{Metrics: }
We report coarse-ranking metric Hit Ratio@$K$ and fine-ranking metric NDCG@$K$~(Normalized Discounted Cumulative Gain), where $K$ is taken as 3,5,10. Moreover, based on the two-step procedure in Performance Ranking, we evaluate performance under two settings: individual queries and overall ranking, distinguished by the subscripts *i* and *o*, respectively. Since our final recommendations are derived from the overall ranking, the individual query metrics are provided for reference only.

\textbf{Tutorial:} If users wants to know the appropriate forecasting models for time series, they just need to submit that dataset as well as set the sampling rate regarding the recommendation speed. ARIES will provide quick feedback on the recommended results as shown in Table~\ref{recom}. Moreover, the users only needs the CPU device because the ARIES recommendation does not involve learnable parameters.

\begin{table*}[t]
	\centering 		
	\caption{The models recommended by ARIES on different datasets and the recommended metrics.}
	\label{recom} 		
	\renewcommand\arraystretch{0.63}
	\resizebox{\linewidth}{!}{
			\begin{tabular}{c|l}
				\toprule[1.5pt]
				\textbf{DataSets} & \textbf{Recommendation Results}\\
				\midrule
				\multirow{21}{*}{\textbf{Electricity}}& \textbf{Interpretability Suggestions:}\\
				& \quad \textbf{Main properties:} 99.58\% Non-stationary \quad 99.13\% Hetro-scedasticity \quad 72.02\% No trend with strength of [0, 0.1]\\
				&\quad 65.03\% Medium-Short-term/ Medium-Low Memory with value  of (0.25, 0.5] \quad 62.39\% Single season\\
				& \quad\textbf{Strategies that can be adopted:} \\
				&\quad Revin, Revin-like, Residual, Channel Interaction, Decomposition(Moving Avg), Season-related, DownSample, \\
				& \quad Multi-scale, Patch, Channel Indepency, Timestamp Embedding, Fourier method\\
				& \quad \textbf{Strategies to be avoided:} \\
				&\quad Time-invariant, Decomposition(Fourier Method), Channel Dependency, Channel Embedding, Foundation Model\\
				&\quad \textbf{Models with potential preferences:} \quad NBeats, UMixer, CATS, DLinear, Crossformer\\
				& \quad \textbf{Potentially unsuitable Models:} \quad\quad\quad NSTransformer, Moirai-Base, DeepAR, Moirai-Large, HI\\ \cline{2-2}
				& \textbf{Top 10 Recommended Models:}\\
				& \quad NBeats, SOFTS, UMixer, MTSMixer, iTransformer, CATS, CycleNet, DSFormer, TimeMixer, Crossformer\\ \cline{2-2}
				& \textbf{Validation:} \\
				& \quad\textbf{The 10 Best Models for Real:}\\
				& \quad CATS, SOFTS, iTransformer, CycleNet, TimeMixer, PatchTST, NBeats, SegRNN, Fredformer, DSFormer\\
				&Hit Ratio@5\_{i}: 0.735  \quad  NDCG@5\_{i}: 0.232 \quad Hit Ratio@5\_{o}: 1.0  \quad  NDCG@5\_{o}: 0.345\\
				&Hit Ratio@7\_{i}: 0.838 \quad  NDCG@7\_{i}: 0.303 \quad Hit Ratio@7\_{o}: 1.0  \quad  NDCG@7\_{o}: 0.744 \\
				&Hit Ratio@10\_{i}: 0.924 \quad  NDCG@10\_{i}: 0.403 \quad Hit Ratio@10\_{o}: 1.0  \quad  NDCG@10\_{o}: 0.732  \\
				\midrule
				
				\multirow{21}{*}{\textbf{PEMS08}}& \textbf{Interpretability Suggestions:}\\
				& \quad\textbf{Main properties:}\quad 100\% Non-stationary \quad 100\% Hetro-scedasticity \quad 95.02\% Multi season\\
				&\quad 70.56\% Medium-low trend with strength of (0.1, 0.5] \quad 69.3\% Medium-low Volatility with value  of (0.4, 0.6]\\
				& \quad\textbf{Strategies that can be adopted:} \\
				&\quad Revin, Revin-like, Residual, Channel Interaction, Decomposition(Moving Avg), Season-related, \\
				&\quad Decomposition(Fourier Method), Timestamp Embedding, Fourier method, Transformer backbone, Channel Indepency\\
				& \quad\textbf{Strategies to be avoided:} \\
				&\quad Time-invariant,  Channel Embedding, Time Series Foundation Model, Channel Dependency\\
				& \quad\textbf{Models with potential preferences:} \quad Triformer, Crossformer, NBeats, Sumba, SOFTS\\
				& \quad\textbf{Potentially unsuitable Models:} \quad\quad\quad Moirai-Large, HI, Moirai-Base, TiDE, NLinear\\ \cline{2-2}
				& \textbf{Top 10 Recommended Models:}\\
				& \quad NBeats, Crossformer, MTSMixer, UMixer, Sumba, SOFTS, DSFormer, CATS, Fredformer, iTransformer\\ \cline{2-2}
				& \textbf{Validation:} \\
				& \quad\textbf{The 10 Best Models for Real:} \\
				& \quad iTransformer, STID, SOFTS, Triformer, TimeMixer, Crossformer, NBeats, Sumba, PatchTST, Fredformer\\
				&Hit Ratio@5\_{i}: 0.789  \quad  NDCG@5\_{i}: 0.289 \quad Hit Ratio@5\_{o}: 0.0  \quad  NDCG@5\_{o}: 0.0\\
				&Hit Ratio@7\_{i}: 0.882 \quad  NDCG@7\_{i}: 0.341 \quad Hit Ratio@7\_{o}: 1.0  \quad  NDCG@7\_{o}: 0.546 \\
				&Hit Ratio@10\_{i}: 0.951 \quad  NDCG@10\_{i}: 0.437 \quad Hit Ratio@10\_{o}: 1.0  \quad  NDCG@10\_{o}: 0.652  \\
				\midrule
				
				\multirow{21}{*}{\textbf{ETTh1}}& \textbf{Interpretability suggestions:}\\
				& \quad\textbf{Main properties:}\quad 100\% Non-stationary \quad 100\% Hetro-scedasticity\\
				& \quad 84.18\%:Medium-Short-term/ Medium-Low Memory with value  of (0.25, 0.5]\\
				&\quad 69.04\%:Multi season \quad 66.82\%:Medium-low Anomaly with value  of (0.05, 0.1]\\
				& \quad\textbf{Strategies that can be adopted:} \\
				&\quad Revin, Residual, Revin-like, Channel Interaction, DownSample, Patch, Decomposition(Moving Avg)\\
				&\quad Season-related, Multi-scale, Channel Indepency, Fourier method\\
				& \quad\textbf{Strategies to be avoided:} \\
				&\quad Time-invariant,  Channel Dependency, Decomposition(Fourier Method)\\
				& \quad\textbf{Models with potential preferences:} \quad Crossformer, NBeats, Sumba, SOFTS, UMixer, \\
				& \quad\textbf{Potentially unsuitable Models:} \quad\quad\quad Informer, NSTransformer, Sumba, Fredformer, NHiTS\\ \cline{2-2}
				& \textbf{Top 10 Recommended Models:}\\
				& \quad NBeats, Crossformer, UMixer, CATS, SOFTS, MTSMixer, Sumba, Triformer, TimeMoE, TimeMixer\\ \cline{2-2}
				& \textbf{Validation:} \\
				& \quad\textbf{The 10 Best Models for Real:}\\
				& \quad SegRNN, NLinear, SparseTSF, CATS, PatchTST, UMixer, TimeMoE, FiLM, TimeMixer, STID\\
				&Hit Ratio@5\_{i}: 0.678  \quad  NDCG@5\_{i}: 0.171 \quad Hit Ratio@5\_{o}: 1.0  \quad  NDCG@5\_{o}: 0.146\\
				&Hit Ratio@7\_{i}: 0.750 \quad  NDCG@7\_{i}: 0.194 \quad Hit Ratio@7\_{o}: 1.0  \quad  NDCG@7\_{o}: 0.256 \\
				&Hit Ratio@10\_{i}: 0.857 \quad  NDCG@10\_{i}: 0.281 \quad Hit Ratio@10\_{o}: 1.0  \quad  NDCG@10\_{o}: 0.335  \\
				
				\bottomrule[1.5pt]
				
			\end{tabular}
		}
	\end{table*}
	
	\subsection{Recommendation Analysis}
	Analyzing the recommended result in Table~\ref{recom}, we can demonstrate that ARIES can assign different recommendation results based on different data properties. 
	
	\textbf{Recommendation Performance:} ARIES successfully retrieves 7, 6, and 4 out of the top-10 real SOTA models across the three datasets, with an overall Hit Ratio@K close to 1, ensuring a strong performance baseline for recommendations. While there are differences in the order of recommendation, this is mainly from hyper-parameter tuning and subtle performance gaps. Notably, the overall NDCG@10 reaches approximately 0.7 for Electricity and PEMS08 datasets, while the Hit Ratio@10 for individual sequences achieves 0.9, which are empirically considered excellent performance for a parameter-free approach. Although the sequence-wise NDCG@K appears relatively low, this primarily results from random variations between sequences.
	
\begin{figure}[t]
		\centering
		\includegraphics[width=0.40\textwidth]{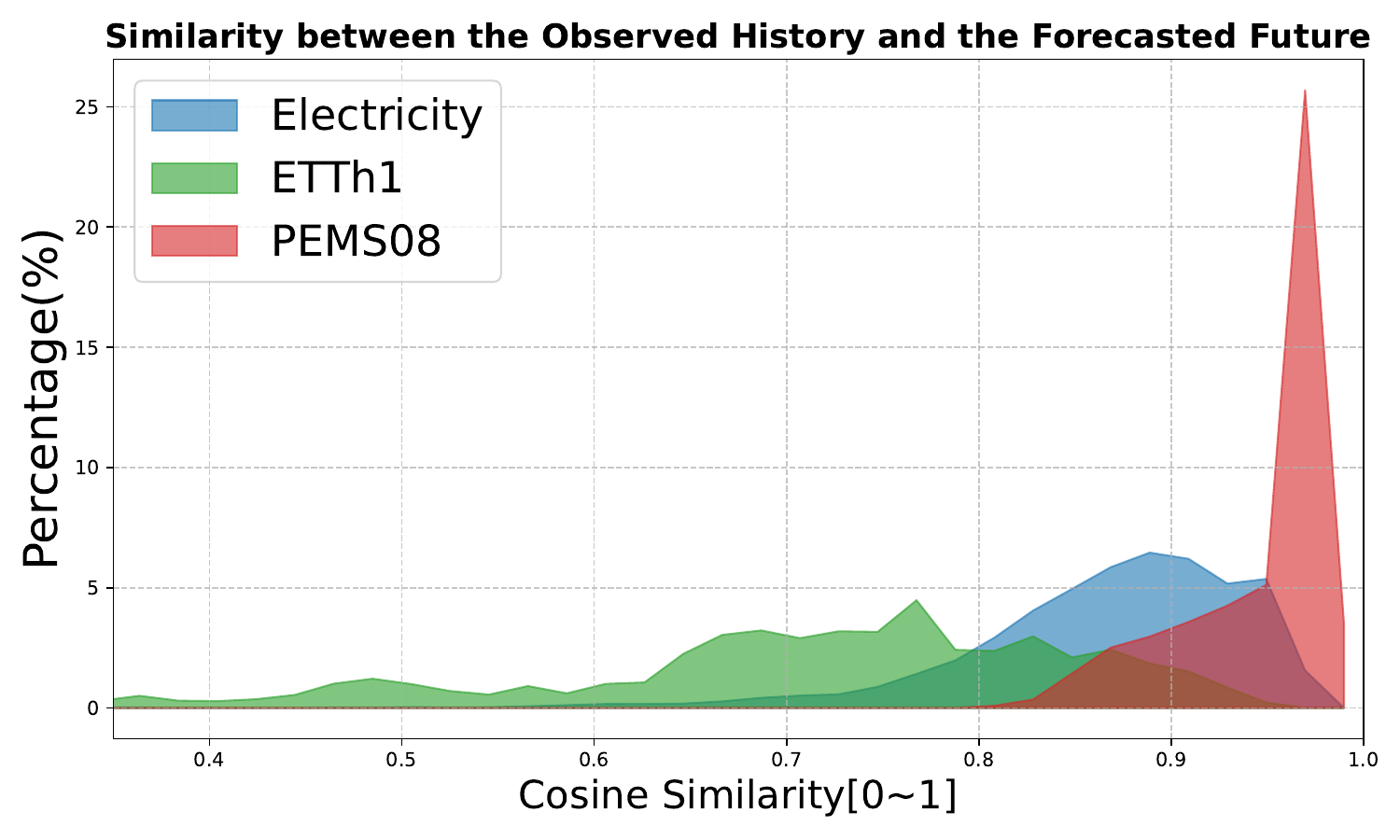}
		\caption{Distribution shift between the observed history and the forecasted future in time series data.}
		\label{simi}
	\end{figure}

	Notably,  recommendation performance on Electricity and PEMS08 significantly outperforms that on ETTh1, which may result in distribution shift between historical and future segments. As illustrated in Figure~\ref{simi}, cosine similarity analysis between historical data and true predictions indicates that ETTh1 exhibits substantially lower similarity compared to Electricity and PEMS08, as detailed in BasicTS~\cite{shao2023exploring}. This instability limits ARIES' recommendation efficacy for ETTh1, which relies on stable temporal pattern continuity.
	
	\textbf{Interpretability suggestion:} After property evaluation, all three benchmark datasets are considered non-stationary, hetero-scedastic and tend to be seasonal with low to medium memory, which leads to convergent partial analyses.  Moreover, the property distribution of benchmark datasets also validate our relation assessments, such as MLP-based methods can match the performance of Transformer-based backbone approaches because the data is inherently less memorable.  Additionally, it also explains the longstanding focus of deep forecasting work on non-stationary and periodic modeling because of the necessity to accommodate significant property preferences.
	
	The suggested strategies cover the most effective approaches for a certain property, and methods such as Revin are often recommended first due to their importance for forecasting. Meanwhile, many of the strategies to be avoided have been validated in previous work such as channel dependency. Models with potential preferences largely conform to the ARIES model recommendation results, and the model suggestions from PEMS08 exactly compensate for the lack of recommendation results. In addition, Potentially unsuitable Models basically excludes the results with very poor performance, except at ETTh1 where inaccurate advice may be given.
	
	\textbf{Efficiency and scalability}: Our parameter-free recommendation method ensures that $Key$-$Value$ and $Query$ construction in Section \ref{sdmpm} and \ref{eppts} are performed independently, making it highly efficient and scalable. Generating recommendations for ETTh1 takes only a few seconds, and for larger datasets like Electricity and PEMS08, the process takes approximately one minute. Meanwhile, users can adjust the sampling rate to significantly accelerate recommendations with minimal impact on performance. Moreover, adding new models or datasets requires only updates to the construction process, without parameter training, resulting in low resource consumption and rapid deployment.
	
	\textbf{Synth only:} Both relation assessment and model recommendation in ARIES are built on diverse patterns of synthetic data. For this reason, we perform additional testing by replacing the Key-Value construction with Benchmark datasets. PEMS08 $\rightarrow$ Electricity: Replacing \textit{Synth} with PEMS08 causes the NDCG@7 and NDCG@10 on Electricity to drop from 0.74 to 0.33, and the number of successful recommendation hits decreases from 7 to 4. Electricity $\rightarrow$ PEMS08: Replacing \textit{Synth} with Electricity decreases NDCG@10  0.65 to 0.53 and NDCG@7 from 0.55 to 0.38 for PEMS08. This is because PEMS08 and Electricity do not have a sufficiently shared model to support their mutual recommendation, and the intrinsic reason for this comes from the differences in their domain rules.
	
	In conclusion, as the first recommendation framework for time series forecasting, ARIES delivers reliable recommendations for stable-pattern datasets. For distributionally shifting data, simpler MLP-based models with complementary preference analysis provide robust and reliable choices,

	\section{Conclusion}
	ARIES not only assesses the relation between time series properties and modeling strategies, but also enables forecasting model recommendation for arbitrary realistic time series. This not only marks a comprehensive and fine-grained analysis of the innovative methods of existing work, but also represents the first realization of deep time series forecasting model recommendation.
	
	Limited by Benchmark's parameter tuning, the mathematical impact of properties and synthetic data, and the assumption of historical-future property consistency, ARIES's modeling analysis struggles to advance much further. In addition to addressing the issues mentioned above, ARIES will continue to maintain its focus on novel work in time series forecasting, enhance our methods for evaluating time series properties, expand the discussions on topics such as spatio-temporal forecasting, time series classification, and so on. Ultimately, we aim to provide a framework for interpretable as well as automated time series analysis to support real-world time series related applications.

\section*{Acknowledgments}
This work is supported by the NSFC underGrant Nos. 62372430 and 62502505, the Youth Innovation Promotion Association CAS No.2023112, the Postdoctoral Fellowship Program of CPSF under Grant Number GZC20251078, the China Postdoctoral Science Foundation No.2025M77154 and HUA Innovation fundings. We thank all the anonymous reviewers who generously contributed their time and efforts.

\bibliographystyle{IEEEtrans}
\normalem
\bibliography{main}
\clearpage

\appendix

\subsection{Property Selection and Stability}~\label{propertyissues}
Notably, our ARIES and BLAST~\cite{shao2025blast}\footnote{https://github.com/GestaltCogTeam/BLAST} frameworks share a unified system for time series properties. BLAST, a parallel our work, focuses on ensuring pattern balance in pretraining corpora for time series foundation models. It evaluates datasets containing over 326B timestamps, applies dimensionality reduction on property results, and performs uniform sampling to achieve balanced pattern distribution—thereby enhancing zero-shot forecasting performance for all foundation models. As a concurrent work, ARIES primarily extends BLAST by enhancing multi-seasonality detection to more accurately quantify time series seasonality and ACF determination for strictly-sense stationarity.

In this section, we explain the criteria and rationale for property selection in ARIES, some of the neglected properties, with a typical pattern display of the selected properties. In addition, we demonstrate the good behavior of the current properties, that is, these properties remain stable when the magnitude, mean, phase and length of time series are changed.

\vspace{10pt}
\subsubsection{Criteria for Property Selection}
The properties selected for ARIES need to support comprehensive and effective time series analysis to support relation assessment and model recommendation. To this end, the following requirements need to be met:
\begin{itemize}
	\item \textbf{Domain Consensus and Mathematical Support:} the selected properties need to have a wide consensus in various application domains and corresponding mathematical computation methods, so as to facilitate the promotion and interpretability of ARIES in downstream applications.
	
	\item \textbf{Good Behavior: }the properties should analyze any real-time pattern, so it should be possible to ensure that they are not affected by pattern-independent factors when dealing with time series with different magnitudes, mean values, and lengths.
	
	\item \textbf{Low Computational Complexity:} ARIES have to support the rapid analysis of large amounts of data to meet the realities of convenience, real-time and other needs, which requires mathematical computation methods need to be kept as low as possible complexity. This is particularly important because property computation is CPU-dependent and greater time complexity is unacceptable, such as when faced with the BLAST 326B data volume.
	
	\item \textbf{Window-independent:} Windows or patches shorter than the observation length are used to analyze subsequence-related properties, but the window needs to manually determine the length as well as high complexity, so ARIES overlook window-related properties.
	
	\item \textbf{History only:} ARIES only analyzes historical observations of a time series and cannot adopt the forecasting component, so properties that analyze the history-future correlation or shift of a time series cannot be included.
\end{itemize}

\vspace{10pt}
\subsubsection{Reasons for Selection and Exclusion}
Rationale for selection of properties in ARIES:
\begin{itemize}
	\item \textbf{Stationarity:} Stationarity is a property that cannot be ignored in the analysis of financial and mathematical time series, and early AR-type methods~\cite{box1968some,hyndman2008forecasting} focused on analyzing time series around this property. The KPSS/ADF test~\cite{dickey1979distribution} and ACF plot methods adopted in our test of stationarity are also the common strategies used in econometrics to determine strictly stationary series. In addition, non-stationarity is a necessary condition to fulfill when exploring time-series patterns, as typical trends and periods are considered non-stationary and have also ignited the exploration of deep time-series forecasting~\cite{wu2021autoformer}.
	
	\item \textbf{Trend \& Season: }Seasonal and Trend decomposition~\cite{cleveland1990stl, bandara2021mstl} is a classical work in mathematical time-series analysis, decomposing the time-series into two simple and mutually exclusive properties. The method is directly inspired by classical mathematical forecasting methods such as ETS and SES~\cite{hyndman2008forecasting}, and those properties are always core motivations for a large amount of work in the field of deep forecasting~\cite{zhou2022fedformer, piao2024fredformer, lin2024cyclenet, zhou2022film, yi2024frequency}. Furthermore, given the decomposition strategy of single-scale convolution~\cite{wu2021autoformer} typically adopted by current methods, we incorporate multi-seasonal considerations~\cite{bandara2021mstl} to explore the limitations of existing work.
	
	\item \textbf{Volatility: }Volatility is an important metric for stochastic processes in mathematical time-series analysis to describe time-series changes, inspiring recent financial forecasting methods~\cite{engle1982autoregressive}. In addition, the relative volatility directly affects the modeling of numerical values in deep forecasting, the volatility space of probabilistic forecasting~\cite{salinas2020deepar, woo2024moirai} and the token strategy of foundational models~\cite{ansari2024chronos}, which supports the future application of ARIES in the broader field of deep time series analysis.
	
	\item \textbf{Memory: }The analysis of long and short-term dependencies has a well-established history in deep learning and is an extremely important motivation in deep temporal forecasting~\cite{zhou2021informer,wu2021autoformer, zhou2022fedformer, nietime, zhou2022film, daslong, linsparsetsf, lin2023segrnn}. In addition, research on backbone selection\cite{zeng2023transformers}, and channeling strategies~\cite{nietime, DBLP:journals/tkde/HanYZ24} also points to the issue of memorability. Distinct from the natural language domain, the memorability of numerical series benefits from the fact that studies in the field of mathematical sciences can be directly measured by the Hurst exponent, although this property is affected by the length of series.
	
	\item \textbf{Scedasticity: }Scedasticity, an important property of econometrics that won the Nobel Prize in Economics in 2003~\cite{engle1982autoregressive}, improves traditional financial forecasting methods, and hetero-scedasticity is extremely common in real time series. In the field of deep forecasting, the same concept of covariance shift has triggered research in time-series transfer learning and distribution shift~\cite{kim2021reversible}.
	
	\item \textbf{Anomaly:} z-score detection is a commonly used strategy in anomaly detection by characterizing outliers that deviate from the normal time series. High anomalies often represent mean shift of values, and together with hetero-scedasticity point to the topic of normalization strategies in Revin~\cite{DBLP:conf/nips/Kim00K24, zhou2022film,wangtimemixer,liuitransformer} and distribution shift~\cite{kim2021reversible}.
	
\end{itemize}

Reasons for excluding some properties:
\begin{itemize}
	\item \textbf{Entropy:}~\cite{aksu2024giftevalbenchmarkgeneraltime} Window inflexibility; High computational costs;
	\item \textbf{Lumpiness:}~\cite{aksu2024giftevalbenchmarkgeneraltime} Duplicates our volatility; Window inflexibility; High computational costs;
	\item \textbf{Shifting:}~\cite{qiu2024tfb} Duplicates our scedasticity; Lack of domain consensus;
	\item \textbf{Transition:}~\cite{qiu2024tfb} Not a numerical form but a matrix; Need for additional definition of modes; Lack of domain consensus;  
	\item \textbf{Correlation:}~\cite{qiu2024tfb} It identifies variable links, but direct computation is too costly for our millions of unaligned subseries. Inspired by the study of channel strategies in \cite{DBLP:journals/tkde/HanYZ24}—a trade-off between parameter capacity and robustness—we generalize it to memory and yield a novel discovery
\end{itemize}

\vspace{10pt}
\subsubsection{Property stability}
Since ARIES needs to face arbitrary time series in real-life scenarios, the property system needs to be able to cope with arbitrary time series, which means that the property computation needs to be as stable as possible in the face of the same time series when the magnitude, mean, phase and length change.

We choose the simple sinusoidal function $sin$ for example and in-depth analysis:

\begin{flalign}
	sin(t) = A * sin(2\pi t /T + \phi) + b, t \in [1, L]
\end{flalign}

where $A$ is the amplitude, $t$ is the time step, $T$ is the period, $\phi$ is the phase, $L$ is the length of series and $b$ is the mean. In the following description, $A$ is usually 1, $T$ is 24, $\phi$ is 0, $L$ is 336 and $b$ is 0.

\begin{itemize}
	\item \textbf{Magnitude change: }The values of the entire time series are multiplied by any real number. The magnitude $A$ changes in the corresponding experiment, taking values between [1, 100] with the slide of 1.
	\item \textbf{Mean change: }The values of the entire time series are added by any real number. The mean $b$ changes, taking values between [0, 100] with the slide of 1.
	\item \textbf{Phase change: }The time series is shifted on the x-axis, and we still observe the full 12 periods due to the setting of seasonal length and time series length. The phase $\phi$ changes, taking values between [0, $180^{\circ}$] with the slide of $10^{\circ}$.
	\item \textbf{Length change: }More timestamps are observed. In strict terms, series of different lengths observe different patterns, so we set the length $L$ to change at [336, 984] with the slide of 24 to observe complete seasons.
\end{itemize}

\textbf{Stationary}: Stationarity calculations are not affected for the above changes and all are judged to be non-stationary.

\textbf{Trend}: For amplitude, mean, and phase changes, the trend strength calculation does not change and is 0.035. When the length changes, the value decreased from 0.035 to 0.012 but does not have an impact on the judgment results.

\textbf{Season}: For amplitude, mean, and phase changes, the season strength calculation does not change and is 0.977. When the length changes, the value decreased from 0.977 to 0.989 but does not have an impact on the judgment results. Regarding season counts, the system counts them as multiple seasons, but more incorrect seasons appears as the length changes.

\textbf{Volatility}: Volatility calculations are not affected for the above changes and all are 0.7071.

\textbf{Memory}: For amplitude, mean, and phase changes, the memory calculation does not change and is 0.289. When the length changes, the value decreased from 0.289 to 0.209, which may have an impact on the judgment of this property.

\textbf{Scadasticity}: Scadasticity calculations are not affected for the above changes and all p-values are close to 0.

\textbf{Anomaly}: Anomaly calculations are not affected for the above changes and all are 0.

In summary, all of the properties behave pretty well. With regard to length changes, trends and seasons make a somewhat minor difference, mainly having an impact on memory, since long and short-term dependence is itself a length-related concept. Since there are no better alternatives for the memorability metrics, the overall results are acceptable, although they would make the application of ARIES somewhat limited.

\subsection{Applicability analysis of synthetic datasets}~\label{synthana}
Gaussian processes can fit the pattern of arbitrary continuous time series, while they cannot fully fit discontinuous points.

\textbf{Non-stationary kernel: }The Exp-Sine-Squared and Dot-Product kernels model non-stationary periodic and trend components, respectively, forming the core of seasonal-trend decomposition methods~\cite{cleveland1990stl, duvenaud2013structure}. Combined with the White-Noise kernel, these Gaussian kernels synthesize time-series patterns akin to those in Moment~\cite{goswami2024moment}, TimeFM~\cite{das2023decoder}, and ForecastPFN~\cite{dooley2023forecastpfn}, covering many real-world scenarios. However, finite Fourier expansions cannot represent arbitrary signals, such as impulse trains with period $T$:

\begin{table*}[t]	 		 		
	\caption{Results of ARIES TEST for two decomposition strategies}
	\renewcommand\arraystretch{1}
	\tiny
	\setlength{\tabcolsep}{1pt}
	\label{tab3}
	\scalebox{1}
	{
\begin{tabular}{|>{\centering}p{40pt}|>{\centering}p{12pt}|>{\centering}p{14pt}>{\centering}p{14pt}|>{\centering}p{14pt}>{\centering}p{14pt}|>{\centering}p{14pt}p{14pt}  |>{\centering}p{14pt}p{14pt}|>{\centering}p{14pt}p{14pt}|>{\centering}p{14pt}p{14pt}|>{\centering}p{14pt}p{14pt}|>{\centering}p{14pt}p{14pt}|>{\centering}p{14pt}p{14pt}|>{\centering}p{14pt}p{14pt}|>{\centering}p{14pt}p{14pt}|>{\centering}p{14pt}p{14pt}|>{\centering}p{14pt}p{14pt}|>{\centering}p{14pt}p{14pt}|
			}
			\hline
			\multicolumn{2}{|c|}{\textbf{Model}} & \multicolumn{2}{c|}{\textbf{Regular}}&\multicolumn{4}{c|}{\textbf{Stationarity}}&\multicolumn{8}{c|}{\textbf{Trend Strength}}&\multicolumn{8}{c}{\textbf{Seasonality Strength}}&\multicolumn{6}{|c|}{\textbf{Seasonality Count}}\\
			
			\multicolumn{2}{|c|}{\textbf{}}&&&\multicolumn{2}{c}{\textbf{Stationary}}&\multicolumn{2}{c|}{\textbf{Non~}}&\multicolumn{2}{c}{\textbf{[0, 0.1]}}&\multicolumn{2}{c}{\textbf{(0.1, 0.5]}}&\multicolumn{2}{c}{\textbf{(0.5, 0.9]}}&\multicolumn{2}{c|}{\textbf{(0.9, 1.0]}}&\multicolumn{2}{c}{\textbf{[0, 0.25]}}&\multicolumn{2}{c}{\textbf{(0.25, 0.5]}}&\multicolumn{2}{c}{\textbf{(0.5, 0.75]}}&\multicolumn{2}{c}{\textbf{(0.75, 1.0]}}&\multicolumn{2}{|c}{\textbf{0}}&\multicolumn{2}{c}{\textbf{1}}&\multicolumn{2}{c|}{\textbf{\(\geq\) 1}}\\
			\hline
			\multicolumn{2}{|c|}{\textbf{}}&Mean&Med.&Mean&Med.&Mean&Med.&Mean&Med.&Mean&Med.&Mean&Med.&Mean&Med.&Mean&Med.&Mean&Med.&Mean&Med.&Mean&Med.&Mean&Med.&Mean&Med. &Mean&Med.\\ 
			\hline
			\multirow{2}{*}{\makecell[l]{\textbf{Autoformer} \\ \textbf{w/ \textit{decom}}}}& MAE&0.499&0.378&0.77&0.799&0.467&0.322&0.376&0.248&0.626&0.527&0.849&0.701&0.394&0.2&0.533&0.378&0.476&0.41&0.488&0.398&0.425&0.282&0.526&0.352&0.426&0.276&0.452&0.344\\
			&MSE&0.597&0.203&1.028&1.001&0.547&0.152&0.355&0.102&0.779&0.397&1.307&0.658&0.507&0.048&0.737&0.179&0.494&0.263&0.544&0.244&0.433&0.126&0.742&0.155&0.454&0.114&0.46&0.189\\
\hline
	\end{tabular}}
\end{table*}

\begin{table*}[t]	 		 		
	\caption{Results of ARIES TEST for RevIN}
	\renewcommand\arraystretch{1}
	\tiny
	\setlength{\tabcolsep}{1pt}
	\label{tab4}
	\scalebox{1}
	{
		\begin{tabular}{|>{\centering}p{40pt}|>{\centering}p{12pt}|>{\centering}p{14pt}>{\centering}p{14pt}|>{\centering}p{14pt}>{\centering}p{14pt}|>{\centering}p{14pt}p{14pt}  |>{\centering}p{14pt}p{14pt}|>{\centering}p{14pt}p{14pt}|>{\centering}p{14pt}p{14pt}|>{\centering}p{14pt}p{14pt}|>{\centering}p{14pt}p{14pt}|>{\centering}p{14pt}p{14pt}|>{\centering}p{14pt}p{14pt}|>{\centering}p{14pt}p{14pt}|>{\centering}p{14pt}p{14pt}|>{\centering}p{14pt}p{14pt}|>{\centering}p{14pt}p{14pt}|
			}
			\hline
			\multicolumn{2}{|c|}{\textbf{Model}} & \multicolumn{2}{c|}{\textbf{Regular}}&\multicolumn{4}{c|}{\textbf{Stationarity}}&\multicolumn{8}{c|}{\textbf{Trend Strength}}&\multicolumn{8}{c}{\textbf{Seasonality Strength}}&\multicolumn{6}{|c|}{\textbf{Seasonality Count}}\\
			
			\multicolumn{2}{|c|}{\textbf{}}&&&\multicolumn{2}{c}{\textbf{Stationary}}&\multicolumn{2}{c|}{\textbf{Non~}}&\multicolumn{2}{c}{\textbf{[0, 0.1]}}&\multicolumn{2}{c}{\textbf{(0.1, 0.5]}}&\multicolumn{2}{c}{\textbf{(0.5, 0.9]}}&\multicolumn{2}{c|}{\textbf{(0.9, 1.0]}}&\multicolumn{2}{c}{\textbf{[0, 0.25]}}&\multicolumn{2}{c}{\textbf{(0.25, 0.5]}}&\multicolumn{2}{c}{\textbf{(0.5, 0.75]}}&\multicolumn{2}{c}{\textbf{(0.75, 1.0]}}&\multicolumn{2}{|c}{\textbf{0}}&\multicolumn{2}{c}{\textbf{1}}&\multicolumn{2}{c|}{\textbf{\(\geq\) 1}}\\
			\hline
			\multicolumn{2}{|c|}{\textbf{}}&Mean&Med.&Mean&Med.&Mean&Med.&Mean&Med.&Mean&Med.&Mean&Med.&Mean&Med.&Mean&Med.&Mean&Med.&Mean&Med.&Mean&Med.&Mean&Med.&Mean&Med. &Mean&Med.\\
			\hline
			\multirow{2}{*}{\makecell[l]{\textbf{Autoformer} \\ \textbf{w/ RevIN}}}& MAE&0.461&0.346&0.773&0.803&0.425&0.302&0.36&0.248&0.572&0.475&0.746&0.673&0.323&0.098&0.451&0.301&0.425&0.371&0.456&0.362&0.405&0.293&0.444&0.269&0.398&0.267&0.43&0.337\\
			&MSE&0.532&0.171&1.042&1.01&0.473&0.131&0.344&0.095&0.648&0.318&1.057&0.615&0.406&0.012&0.584&0.117&0.429&0.2&0.526&0.205&0.401&0.129&0.589&0.089&0.407&0.109&0.43&0.171\\
\hline

			\hline
	\end{tabular}}
\end{table*}

\textbf{Theorem 1 (Gibbs Phenomenon)}:~\cite{gibbs1899fourier}
For a Dirac comb \(\text{III}_T(t)\) of period \( T \), its \(N\mbox{-} times\) truncated Fourier series approximation:
\[
\text{III}_T^{(N)}(t) = \frac{1}{T} \left[ 1 + 2 \sum_{k=1}^N \cos(k \omega_0 t) \right], \quad \omega_0 = \frac{2\pi}{T} \quad k \in N
\]
exhibits an unbounded $L^2$ error:
\[
\epsilon_N = \|\text{III}_T(t) - \text{III}_T^{(N)}(t)\|_2^2 = \sum_{k=N+1}^\infty \left|\frac{1}{T}\right|^2 \propto \sum_{k=N+1}^\infty 1 \to \infty \quad
\]

Thus, discontinuous periods evade precise Fourier-based or sin functions fitting.

\textbf{Stationary kernel: }For stationary kernels (e.g., Rational Quadratic, RBF, Matérn), we formalize their capabilities and limits:

\textbf{Theorem 2 (Generalized Stone–Weierstrass Approximation Theorem)}~\cite{stone1948generalized}: If a kernel \(k(x, x')\) universal (e.g., Rational Quadratic, RBF, Matérn) over a compact domain \(X\) l, then its Reproducing kernel Hilbert space~(RKHS) is dense in the space of continuous functions \(C(X)\). That is, for any continuous function \(f \in C(X)\) and \(\epsilon > 0\), there exists \(h\) in the RKHS satisfying:
\[
\sup_{x \in X} |f(x) - h(x)| < \epsilon
\]
Hence, Gaussian processes can approximate arbitrary continuous functions.

\textbf{Theorem 3 (Unapproximability of discontinuous functions)}: Consider the discontinuous step function \(f(x) = \mathbb{I}_{x \geq x_0}\), \(f(x)\) has a left limit of \(0\) at \(x_0\) and a right limit of \(1\), so there exists a neighborhood \((x_0 - \delta, x_0 + \delta)\) such that:
\[
|h(x) - f(x)| \geq \frac{1}{2}, \quad \exists x \in (x_0 - \delta, x_0 + \delta)
\]
This contradicts Theorem 2, proving stationary kernels cannot fit discontinuities.

\textbf{More explanation}: Gaussian processes tend to synthesize only continuous, significant time-series patterns. Discontinuities in series generated by ARIES can only come from the White-Noise kernel which violates the Lipschitz condition, and given the research position of ARIES, mutations due to missing data are not considered.

\subsection{Strategy Analysis Demonstration}~\label{strategyana}
To demonstrate the generalizability of ARIES TEST for a wide range of deep forecasting analysis, as well as to expose deeper strategy-property relations, we select two important modeling strategies for extensive ablation studies to encourage more robust and effective innovations in this field.

We conduct ablation studies on the time series decomposition strategy and RevIN. In the reported tables, green indicates performance degradation of the component, red denotes improvement below $Regular's$ level, while blue represents improvement exceeding $Regular's$ performance.

\vspace{10pt}
\subsubsection{Seasonal Trend Decomposition}
\textbf{Decomposition strategy:} Moving Average method involves extracting and separating the trend information of a time series by a convolutional kernel of all 1s. Fourier-based method refers to separating the seasonal information of the magnitude spectrum TopK:

\begin{itemize}
	\item \textit{Moving Average: } Autoformer; DLinear; PatchTST; TimeMixer
	\item \textit{Fourier-based: } ETSformer; Koopa; TimeMixer
\end{itemize}

\textbf{Experimental setup:}  We conduct ablation studies for each decomposition strategy of the baseline, with all parameters remaining consistent throughout. Additionally, since TimeMixer offers both decomposition options, we performed experiments for both scenarios.

\textbf{Findings and analysis: }Observing the results in Table~\ref{tab3}, both decomposition strategies generally improve performance but still impair some properties, and the same strategy exhibits consistent performance gains. Surprisingly, \textbf{Moving Average and Fourier-based methods exhibit completely opposite preferences} in terms of trend, seasonal strength and count, memory, and anomaly, which is more pronounced in the two variants of TimeMixer. Moving average enhances forecasting performance in strongly seasonal scenarios, whereas the Fourier-based method improves strongly trended ones, implying that decomposing a particular pattern first may harm its forecasting capability.

We hypothesize that precisely extracted patterns often struggle to persist in a highly stochastic future, whereas disturbance term removal improves the perception of another pattern. In addition, previous work has argued that Moving average, Fourier methods give a general boost, without knowing their opposite preferences, mainly because benchmark datasets lack the pronounced patterns seen in \textit{Synth}. Most real-world data are superimposed trends and seasons, exhibiting moderate trend and seasonal strength.

\vspace{10pt}
\subsubsection{Reversible Instance Normalization}
\textbf{RevIN:} Such methods often serve as model-agnostic plugins to mitigate distribution shifts in time series data. Nowadays, most models typically incorporate a simple RevIN (Reversible Instance Normalization) module to ensure a basic performance lower bound. 

\textbf{Experimental setup:} We select a number of models for the ablation study of RevIN and the parameters of the models are consistent.

\textbf{Findings and analysis: }The improvement brought by RevIN to the model is universal and substantial, serving as the core guarantee for the performance of iTransformer, DSformer, and TimeNet. Moreover, RevIN's enhancement behavior is consistent, particularly for prominent patterns such as strong seasonality and long-term dependencies.

Strictly speaking, patterns and distributions are two distinct properties. Given the pervasiveness of distribution shifts, separating distributional information can significantly simplify pattern learning. Therefore, RevIN ensures a lower bound for forecasting performance and is a key reason why modern MLP-based models can approach the performance of Transformer-based ones. We also note that distribution shift is an independent topic in time series analysis, often leading to model-agnostic plugin innovations. We advocate that any specific model innovation must include—and should be limited to—a simple implementation of RevIN to ensure fairness in comparisons with other models.

\vspace{10pt}
\subsubsection{Experimental Environment}
Our experiments fully adopt the default configuration of the public-source and fair benchmark BasicTS
, including ADAM as the default optimizer, with each model having its own dedicated parameter configuration file for each dataset. All experiments are conducted on a single NVIDIA GeForce RTX 4090 GPU, with an Intel(R) Xeon(R) Gold 6338 CPU @ 2.00GHz, and each forecasting experiment is limited to 4 threads. In addition, the computation of time series characteristics requires a full CPU thread and does not involve the GPU.

Therefore, if researchers wish to utilize ARIES TEST to evaluate the performance of their forecasting model and strategy preferences, they need GPU resources as well as the framework of BasicTS; if they wish to recommend appropriate models and strategies for the real-world data, they need to guarantee at least 16 threads of CPU to accelerate the property computation. More code information will continue to be updated on our GitHub link: \url{https://github.com/blisky-li/ARIES}.

\subsection{Model Recommendation Algorithm}\label{code}
Algorithm~\ref{alg:recommendation} is the pseudo-code for model recommendation in ARIES, which helps readers understand this parameter-free collaborative filtering method. In summary, it uses time series property as the bridge to query the most similar \textit{Synth}'s subseries in the real-world dataset. Based on the queried performance and the corresponding results of the property-strategy, it provides downstream applications with appropriate forecasting models and interpretable recommendation reasons.

\begin{algorithm*}[!ht]
	\setstretch{1}
	\renewcommand{\algorithmicrequire}{\textbf{Input:}}
	\renewcommand{\algorithmicensure}{\textbf{Output:}}
	\caption{Model Recommendation via Property-Based Retrieval}
	\label{alg:recommendation}
	
	\begin{algorithmic}[1]
		\Require
		\Statex Synthetic dataset $\mathcal{S} \in \mathbb{R}^{B \times N \times L}$ ~(system-included); 
		\Statex Model Performances $\mathcal{M}$~(system-included);
		\Statex Binning strategy $\mathcal{B}$ (from Fig.~\ref{datasetsbing}, system-included);
		\Statex Practical dataset $\mathcal{Q} \in \mathbb{R}^{B' \times N' \times L'}$~(user-input);
		\Statex Proportion of sample retrieval $\tau$~(which controls speed, user-input)
		\Ensure Recommended model list $\mathcal{R}$;
		
		\State Initialize key-value store $\mathcal{KV} \gets \emptyset$
		\For {$i = 1$ \textbf{to} $B$} 
		\For {$j = 1$ \textbf{to} $N$}
		\State $K_{i,j} \gets \mathcal{S}[i,j,1:L]$ \Comment{Initial Construction of Keys}
		\State $V_{i,j} \gets \{(\mathcal{M}_m, \text{MAE}_{i,j}^m, \text{MSE}_{i,j}^m)\}_{m=1}^M$ \Comment{Value Construction}
		\State $\mathcal{KV} \gets \mathcal{KV} \cup \{(K_{i,j}, V_{i,j})\}$
		\EndFor
		\EndFor
		
		\State  Build property index $\mathcal{I} \gets \emptyset$
		\For {each $(K, V) \in \mathcal{KV}$} 
		\State $\mathbf{p} \gets \mathcal{B}(K)$  \Comment{Key Compression}
		\State $\mathcal{I}[\mathbf{p}] \gets \mathcal{I}[\mathbf{p}] \cup \{(K, V)\}$
		\EndFor
		
		\State Initialize query group counter: $\mathcal{G} \gets \emptyset$
		\For {$i' = 1$ \textbf{to} $B'$} 
		\For {$j' = 1$ \textbf{to} $N'$}
		\State $\mathcal{G}[\mathbf{p}_{i'j'}] \gets \mathcal{B}(\mathcal{Q}[i',j',1:L'])$  \Comment{Query Grouping}
		\EndFor
		\EndFor
		
		\For {each unique $\mathbf{p}' \in \mathcal{G}$}
		\State Retrieve: $\mathcal{C} \gets \mathcal{I}[\mathbf{p}']$ 
		\If{$\mathcal{C} = \emptyset$}
		\State Find nearest: $\mathcal{C} \gets \arg\min_{\mathbf{p}_k \in \mathcal{I}} \|\mathbf{p}_k - \mathbf{p}'\|_1$  \Comment{Vector Retrieval}
		\EndIf
		\State Sample values: $\tilde{V} \gets \text{WeightedSample}(\mathcal{C}, \tau \times |\mathcal{G}[\mathbf{p}']|)$   \Comment{Performance Sampling}
		\EndFor

		\State Initialize score table $\mathcal{T} \gets \{\}$
		\For {each sampled $\tilde{v}' \in \tilde{V}$}
		\For {$m = 1$ \textbf{to} $M$} 
		\State $\mathcal{T}[m] \gets \mathcal{T}[m] \cup \{\text{MAE}_m^{\tilde{v}'}, \text{MSE}_m^{\tilde{v}'}\}$ \Comment{Individual Recording}
		\EndFor
		\EndFor
		\State $\mathcal{R} \gets \text{argsort}_m \left( \frac{1}{|\mathcal{T}[m]|} \sum \mathcal{M}_m^{\mathcal{T}} \right)$ \Comment{Overall Ranking}
	\end{algorithmic}
\end{algorithm*}

\subsection{Discussion}\label{discussion}
\vspace{10pt}
\subsubsection{Scope of ARIES}
The scope of ARIES is consistent with the time series forecasting task. The time series data in this paper are structured sequences of numerical type, unstructured and requiring additional processing such as text, video, speech, and trajectory sequences are usually not considered in this scope.

In addition, considering our task setup, the mathematical limitations of synthetic data and the performance boundaries of the recommender system, ARIES is mainly suitable for the analysis and recommendation of remarkable temporal patterns. Highly noisy and distributionally shifting data are currently difficult to analyze effectively due to the assumption of historical-future property consistency, which is our main subsequent improvement direction.

\vspace{10pt}
\subsubsection{Property Selection}
The selected properties for ARIES must fulfill five criteria: domain-agnostic mathematical foundations for interpretability, robustness to variations in scale/mean/length, low computational complexity for real-time analysis, window-independent pattern extraction (avoiding manual segmentation), and exclusive reliance on historical observations while excluding forecasting components.

With reference to existing work, we exclude Entropy and Lumpiness~\cite{aksu2024giftevalbenchmarkgeneraltime} due to manual sliding window, Shifting and Transition~\cite{qiu2024tfb} because of the poorer math interpretability. In addition, while Correlation is considered important for measuring variable relationships, we incorporate a discussion of channeling into memorability due to issues of computational complexity as well as inspiration from existing work~\cite{DBLP:journals/tkde/HanYZ24}.

\vspace{10pt}
\subsubsection{Property Applicability}
To enable the adoption of ARIES properties for measuring arbitrary time series, the property computations must remain robust against translation~(mean shifts), scaling~(amplitude variations), and warping~(length variations).

We conduct systematic tests and observe that mean and amplitude variations do not affect property evaluations. While warping introduces negligible impacts on trend and seasonal strength calculations, it induces non-linear effects on memory property estimation. This occurs because temporal memory inherently depends on the ratio of periodicity to the overall sequence length. Considering this problem, we suggest that the length of time series processed by ARIES be between 96 and 2024, which is the usual setting for time series forecasting.

\subsection{Future Work}

ARIES is envisioned as a starting point for deep forecasting technologies. Our goal is to explore the underlying principles of forecasting tasks and promote their practical deployment across diverse real-world domains.

In the short term, we plan to expand ARIES with a more comprehensive time series characterization system, broader evaluations of forecasting models and strategies, and a parameter-free, interpretable model recommendation algorithm with stronger performance.

Looking ahead, ARIES will integrate with efforts such as BasicTS and BLAST, incorporating emerging techniques like automated hyper-parameter tuning, time series forecast enhancement technology, and temporal pattern understanding. Our long-term vision is to offer an assembly-style modeling framework that directly delivers optimal forecasting pipelines, which completely eliminates technical barriers for downstream applications.

\vfill

\end{document}